\documentclass[journal]{IEEEtran}
\usepackage{graphicx}
\usepackage{amsmath}
\usepackage{makecell}
\usepackage{multirow}
\usepackage{collcell}
\usepackage{colortbl}
\usepackage{bm}
\usepackage{cite}
\usepackage{tikz}
\usepackage{pgfplots}
\usepackage{pgfplotstable}
\usepackage{epsfig}
\usepackage{enumitem}
\usepackage{adjustbox}
\usepackage{soul}
\usepackage{newfloat}
\usepackage{xcolor}
\usepackage{siunitx}
\usepackage{collcell}
\usepackage{graphicx} 
\usepackage{tikz}
\usepackage[skip=-1.85mm]{subcaption}
\usepackage{pgfplots}
\usepackage{mathtools}
\usepackage{multirow}
\usepackage{url}
\usepackage{booktabs}
\usepackage[font=small]{caption}

\usepackage[T1]{fontenc}  
\usepackage{amsmath,amsfonts,amssymb,amsthm}
\DeclareFloatingEnvironment[fileext=dia, within=section]{diagram}
\usepackage[normalem]{ulem}
\useunder{\uline}{\ul}{}
\usepackage{tkz-kiviat}
\usetikzlibrary{arrows}
\usetikzlibrary{patterns}

\newcommand{\TrainingSet}{\bm{T}}
\newcommand{\ValidationSet}{\bm{V}}
\newcommand{\TestSet}{\Psi}

\usepackage{cleveref}

\usetikzlibrary{decorations.pathmorphing}
\usetikzlibrary{arrows,positioning}
\usetikzlibrary{plotmarks}
\usetikzlibrary{calc}
\usetikzlibrary{shapes.geometric}
\usetikzlibrary{decorations.pathreplacing,calligraphy}
\usetikzlibrary{arrows,patterns}
\usepgfplotslibrary{groupplots}
\usetikzlibrary{patterns}
\pgfdeclarelayer{bg}    
\pgfsetlayers{bg,main, background}
\usepgfplotslibrary{fillbetween}
\pgfplotsset{compat=1.17, compat/show suggested version=false}
\def\Decimal{0.0000}

\def\Ulinehelp#1.#2 {%
  #1.#2\setbox0=\hbox{#1\Decimal}\hspace{-\wd0}{\if\relax#2\relax%
    \uline{\phantom{#1.0}}\else\uline{\phantom{#1.#2}}\fi}%
}
\usetikzlibrary{patterns}

\usetikzlibrary{spy}
\usepackage{algorithm}
\usepackage{algpseudocode}

\ifCLASSINFOpdf
\else
\fi

\definecolor{hotmagenta}{rgb}{1.0, 0.11, 0.81}
\definecolor{gray1}{RGB}{0,0,0}
\definecolor{gray2}{RGB}{80,80,80}
\definecolor{gray3}{RGB}{140,140,140}
\definecolor{gray4}{RGB}{170,170,170}
\definecolor{gray5}{RGB}{200,200,200}
\definecolor{gray6}{RGB}{220,220,220}
\definecolor{coordscolor}{RGB}{235,235,235}

\begin{document}
\title{Graph Neural Networks Extract High-Resolution Cultivated Land Maps from Sentinel-2 Image Series}
\author{
Lukasz Tulczyjew, Michal Kawulok,~\IEEEmembership{Member,~IEEE}, Nicolas Long\'{e}p\'{e}, Bertrand Le Saux,~\IEEEmembership{Senior Member,~IEEE}, and Jakub Nalepa,~\IEEEmembership{Member,~IEEE}
\thanks{LT, MK and JN are with Silesian University of Technology, Gliwice, Poland (e-mail: jnalepa@ieee.org) and with KP Labs, Gliwice, Poland. NL and BLS are with $\Phi$-lab, European Space Agency, Frascati, Italy.\\This work was funded by the European Space Agency (the GENESIS project), and supported by the ESA $\Phi$-lab ({https://philab.phi.esa.int/}). LT, MK and JN were supported by National Science Centre, Poland
(2019/35/B/ST6/03006).}
}

\markboth{IEEE GEOSCIENCE AND REMOTE SENSING LETTERS}%
{Shell \MakeLowercase{\textit{et al.}}: Bare Demo of IEEEtran.cls for IEEE Journals} 

\maketitle
 
\begin{abstract}

Maintaining farm sustainability through optimizing the agricultural management practices helps build more planet-friendly environment. The emerging satellite missions can acquire multi- and hyperspectral imagery which captures more detailed spectral information concerning the scanned area, hence allows us to benefit from subtle spectral features during the analysis process in agricultural applications. We introduce an approach for extracting 2.5\,m cultivated land maps from 10\,m Sentinel-2 multispectral image series which benefits from a compact graph convolutional neural network. The experiments indicate that our models not only outperform classical and deep machine learning techniques through delivering higher-quality segmentation maps, but also dramatically reduce the memory footprint when compared to U-Nets (almost 8k trainable parameters of our models, with up to 31M parameters of U-Nets). Such memory frugality is pivotal in the missions which allow us to uplink a model to the AI-powered satellite once it is in orbit, as sending large nets is impossible due to the time constraints.


\end{abstract}


\IEEEpeerreviewmaketitle

\section{Introduction} \label{sec:intro}

Maintaining farm sustainability by improving the agricultural management practices has become an important issue nowadays to ensure sustainable food security~\cite{10.3389/fsufs.2020.00098}, as appropriate management of the cultivated land is paramount for sustaining economic growth~\cite{kpienbaareh2021crop}, and can help us deal with the climate change. As a consequence of the rapid growth of the Earth observation (EO) satellites~\cite{9606737}, we are currently able to acquire image data capturing detailed spectral information over large areas, e.g.,~through the Copernicus Programme, where the Sentinel-2 (S-2) Multi-Spectral Images (MSIs) can be freely accessed. Therefore, the recent advances in EO and artificial intelligence (AI) can play a pivotal role in scalable monitoring of large areas which would not be feasible through the time-consuming and costly in-situ measurements. The S-2 MSIs embrace 13 bands of varying spatial resolution (60\,m, 20\,m, and 10\,m GSD)---extracting information from such imagery has been important in an array of tasks, including environmental monitoring~\cite{grabska2019forest}, change and object detection, estimating crop production or vegetation analysis thanks to the available spectral information~\cite{10.3389/fenvs.2020.00085}. Such data is, however, highly-dimensional and may be redundant, and extracting the relevant information from it remains an open issue. It is especially important for on-board AI in EO, because it can allow us to prune unnecessary data before its transfer. On the other hand, the highest available 10\,m GSD bands may still be a limiting factor in some use cases, such as detecting or measuring small land areas. To tackle this issue, we may benefit from single- and multi-image super-resolution (SR) reconstruction algorithms which enhance the spatial image resolution~\cite{9627135}. However, this process should not impact the spectral information~\cite{LANARAS2018305}. The SR methods range from classical interpolations~\cite{10.7717/peerj-cs.621} to data-driven models which learn the correspondence between the low- and high-resolution data~\cite{8884136}. Unfortunately, existing SR solutions are application-agnostic and they are seldom validated for a specific EO task. Finally, MSIs can be acquired for the same area in multiple time points---this temporal aspect opens doors for other use cases, such as monitoring of vegetation or natural disasters~\cite{bioresita2019fusion}.

\begin{figure}[t!]
\centering
\includegraphics[width=1\columnwidth]{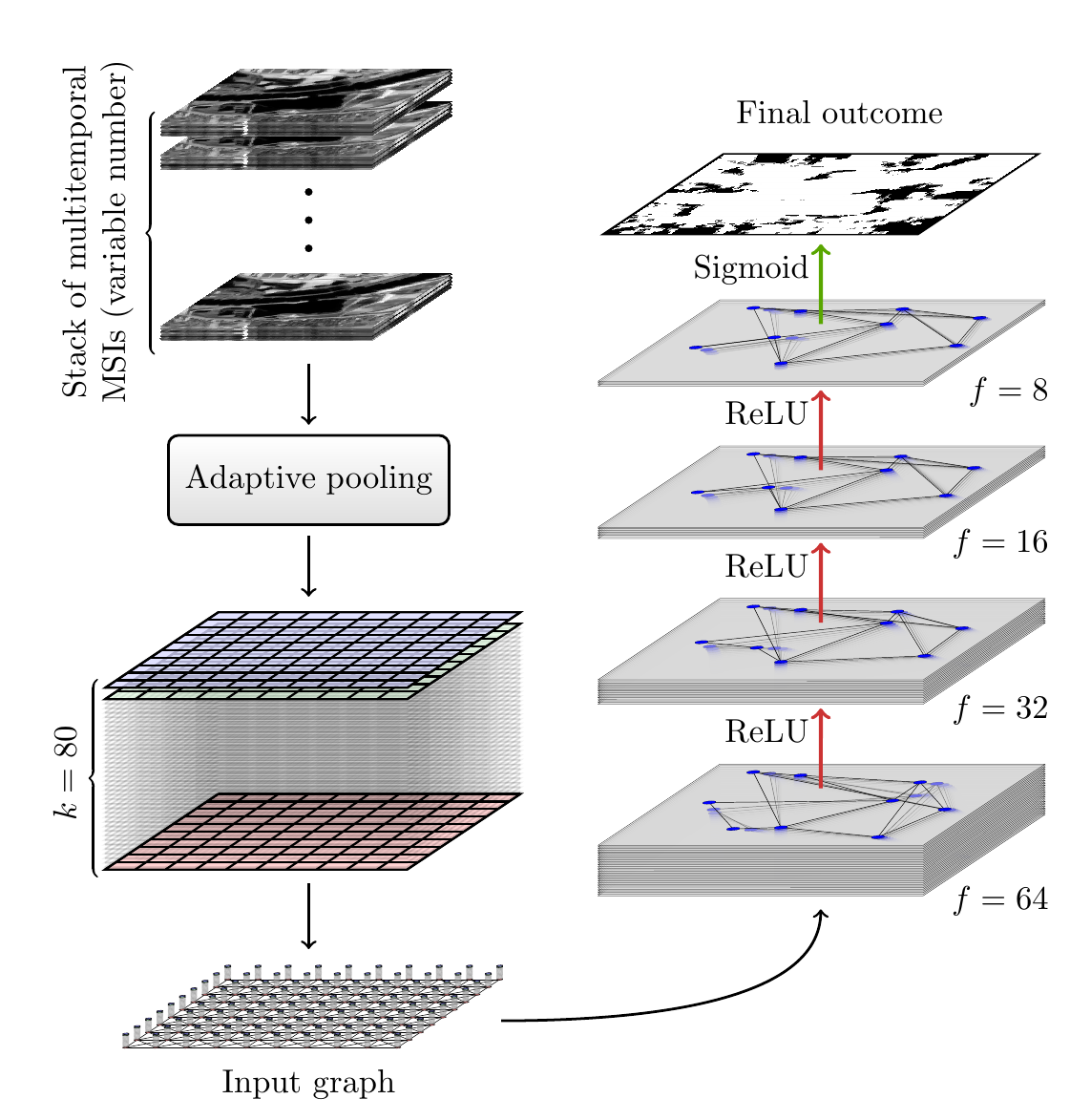}
\caption{Graphical representation of the proposed GCNN for extracting high-resolution cultivated land maps from S-2 image series---as the output, we return the segmented cultivated land map.}\label{fig:gcnn}
\end{figure}

There is a need for bringing AI into space to automate image analysis~\cite{9600851}. This allows us to not only ``keep the brain next to the eyes'', but also to minimize the amount of data to transfer and accelerate the response time to the monitored events. While various architectures were exploited for processing such imagery~\cite{8738045}, convolutional neural nets (CNNs) are dominant, as they can capture spectral and spatial features. Although CNNs can be optimized for the on-board use~\cite{NALEPA2020102994}, they can still be too large to be efficiently transferred to a reconfigurable satellite. Recently, graph convolutional neural nets (GCNNs) that can be resource-frugal and are applicable to irregular data~\cite{ZHOU202057} started gaining attention~\cite{9506070}. We follow this research avenue to build lighter yet performant segmentation models.

We introduce an approach for extracting cultivated land maps from S-2 MSIs (Fig.~\ref{fig:gcnn}). We address the challenges of (\textit{i})~building the algorithms for accurate segmentation of multitemporal MSIs that can generalize well over the unseen data, and (\textit{ii})~developing compact deep learning models (Section~\ref{sec:dataset_and_methods}). The experiments performed over the real-life data released within the framework of the Enhanced S-2 Agriculture Challenge\footnote{See details at the permanently-open challenge page: \url{https://platform.ai4eo.eu/enhanced-sentinel2-agriculture-permanent} (accessed on Mar. 24, 2022).} revealed that our approach outperforms classical and deep learning methods and delivers higher-quality cultivated land maps (Section~\ref{sec:algorithm}). Also, we dramatically reduced the memory requirements of our models (quantified as the number of trainable parameters which was reduced up to $3900\times$ when compared to U-Nets) while obtaining better segmentation, and showed that the GCNN offers fast inference. We made our framework publicly available at \url{https://gitlab.com/jnalepa/gcnn4sentinel2} to maintain full reproducibility.

\section{Materials and Methods}\label{sec:dataset_and_methods}
In this section, we discuss the dataset, together with the training/test split exploited in the experimental study (Section~\ref{sec:dataset}). Our GCNNs for the cultivated land segmentation from S-2 image series are presented in Section~\ref{sec:algorithm}.

\subsection{Dataset Description}\label{sec:dataset}

The S-2 MSI time-series data was acquired for the growing season (March--September 2019) in the Republic of Slovenia and its neighboring countries within the Enhanced S-2 Agriculture Challenge. The original area is divided into 125 scenes, 100 of which are labeled, whereas for the remaining 25 scenes the labels have not been disclosed. Each scene covers an area of $5\times5$~km, forming a $500\times500$ pixels region of interest (at the 10\,m GSD) that is paired with the $2000\times2000$ ground-truth (GT) cultivated land map of a higher resolution with the 2.5\,m GSD. Here, the challenge organizers have upscaled all lower-resolution bands (of 20\,m and 60\,m GSD) to 10\,m in the input image stacks. Therefore, the task is to not only segment the cultivated land map from the 12-band S-2 time series (ranging from 19 to 48 geospatially co-registered MSIs; band B10 was removed by the organizers), but also to upsample it to 2.5\,m GSD at the same time. The 2.5\,m GSD of super-resolved S-2 images was shown to be sufficient to allow for accurate geometrical analysis of small objects and finer descriptions and change detections in many areas, including agriculture~\cite{rs12152366}.

To quantitatively assess the segmentation performance, we focus on the 100 scenes for which the GT is known. We randomly sampled 20 scenes and included them in the test set $\TestSet$, while the remaining 80 scenes form the training set $\TrainingSet$. It is worth mentioning that a similar split was used by our team internally during the challenge to locally assess the quality of our techniques, and the relations across the investigated techniques were fairly consistent with those for the 25 scenes with undisclosed GT that were assessed on the server.

\subsection{Proposed Method}\label{sec:algorithm}
Our proposed approach is depicted in Fig.~\ref{fig:gcnn}, and it can be decomposed into several steps that are performed sequentially. First, the MSIs from an input time series are stacked along the spectral axis, resulting in a variable-length collection of MSIs. Subsequently, the concatenated input is forwarded into an adaptive max pooling layer, which extracts a constant number of temporal and spectral features for each pixel in the scene. This operation allows us to maintain a constant number of fused bands, hence the size of feature vectors for each node in the GCNN is constant as well. Afterwards, the resulting tensor is upsampled utilizing the bicubic interpolation to match the target size (here, the 2.5\,m cultivated land map). The GCNN incorporates four hidden layers, consisting of $64$, $32$, $16$ and $8$ activations per node, respectively, and the output layer with a single unit (in Fig.~\ref{fig:gcnn}, we denote this hyperparameter as $f$). The activation function employed in all hidden layers is the rectified linear unit (ReLU), whereas sigmoid is used in the output layer. Finally, the probability map is thresholded (we use the threshold of 0.5) to obtain the cultivated land maps.


Each node in the graph that becomes the input to GCNN represents a different pixel in the scene (we utilize the 8-connectivity). Our GCNN is a single-parameter model, with only one weight matrix in each layer. Such strategy allows for minimizing the effect of overfitting for graphs with a limited number of labeled nodes. In the cultivated land segmentation task, each node of the input image is labeled (cultivated land vs. background). However, due to the high memory requirements concerned with processing the entire scene at a time, we split it into non-overlapping patches. Therefore, utilizing a single-parameter model should address the problem of overfitting to such spatially-reduced samples. 

Each $(l+1)$-\textit{th} layer in GCNN can be given as~\cite{kipf2016semi}:
\begin{equation}
{
    {H^{l+1}}={\rm ReLU}({\hat{A}}{H^l}{W^l}),
}\label{eq:gcnn_layer}
\end{equation}
where ${H^l}$ and ${W^l}$ represent the activation and learnable weight matrices of layer $l$, respectively. In the first layer of the model ($l=1$), ${H^l}$ constitutes the input patch in matrix-based format, where the number of rows is equal to the number of pixels, and columns define the feature vectors of each node. Furthermore, ${\hat{A}}$ is the normalized adjacency matrix:
\begin{equation}
{
    {\hat{A}}={D^{-\frac{1}{2}}}{A}{D^{-\frac{1}{2}}},
}\label{eq:adj_matrix}
\end{equation}
where ${D}$ is the diagonal {degree matrix} calculated by summing over all columns of ${A}$ in each consecutive row $i$, and emplacing the value in ${d_{i,i}}$. We incorporate the self-connections within the graph, hence each node can utilize its own features during the aggregation step. It is achieved by adding the identity matrix to the adjacency ${\widetilde{A}}={A}+{I}$, and recomputing the degree matrix from ${\widetilde{A}}$. Consequently, we have:
\begin{equation}
{
    {\hat{A}}={\widetilde{D}^{-\frac{1}{2}}}{\widetilde{A}}{\widetilde{D}^{-\frac{1}{2}}}.
}\label{eq:adj_matrix_with_self_connections}    
\end{equation}



\noindent To control the magnitudes of the activations aggregated for each node, we perform the adjacency matrix normalization. This procedure becomes extremely useful when a vertex shares a lot of connections with other nodes. Such phenomenon would induce large values in their representations, while maintaining lower activations for the ``border'' vertices, which consequently may lead to the exploding and vanishing gradient problems~\cite{hanin2018neural}. When the adjacency matrix is normalized, the representation of each node is calculated as a weighted mean of its neighbors (including itself), which additionally helps improve stability of the training phase. Finally, it is worth mentioning that in the $(l+1)$-\textit{th} layer we maintain the $(l+1)$ order of relationship~\cite{DBLP:conf/icml/Le0V20}. It means that e.g.,~in the second layer, the aggregation of features for a  node covers not only its neighbors but also the vertices connected to them. It allows us to enhance the contextual information captured by GCNN, hence to detect more abstract spatial and temporal features.

\section{Experimental results}
\label{sec:experiments}

The main objective of our study is to investigate the segmentation capabilities of the proposed GCNN, and to confront it with both the deep learning and classical machine learning techniques for the task of segmenting cultivated land from S-2 image series. For comparison, we took a U-Net model which established the state of the art in a wide range of image segmentation tasks~\cite{ronneberger2015u}, together with a recent long short-term memory (LSTM) network for segmenting S-2 image stacks treated as time series~\cite{rs13142790}, and a random forest (RF) classifier which utilizes both spectral and spatial features, averaged across the temporal dimension (the MSIs are co-registered). The U-Net follows the topology introduced in~\cite{ronneberger2015u}, and we apply batch normalization after each $3\times3$ convolutional layer in the contracting and expansive paths. Furthermore, we evaluate two versions of this architecture: in U-Net-B, we exploit the bicubic interpolation to spatially upsample the feature maps in the contracting pass. In U-Net-TC, the transposed convolutional layers are used to tackle the intrinsic super-resolution task. For RF, we clean the data by applying the cloud masks for each image. Afterwards, for each pixel, we calculate its spatial and spectral statistical features by investigating its $5\times5$ neighboring patch (the pixel of interest is the central one). Here, the extraction process is performed within the image stack, and we obtain the minimum, maximum, mean, median, standard deviation, and the 1$^{\rm st}$, and 3$^{\rm rd}$ quartile of the pixel values, together with the span of values within the patch. The features are extracted for each band separately, therefore the size of the feature vector for each patch is $96=8\cdot 12$, since there are 8 statistical properties and 12 S-2 bands. This approach allowed us to take the 6$^{\rm th}$ place (out of 17 teams) in the Enhanced S-2 Agriculture Challenge.

The investigated approaches were implemented in \texttt{Python 3.8} with \texttt{Pytorch 1.10.0} and \texttt{PyTorch Geometric}. The experiments were run using an NVIDIA Tesla T4 GPU (16~GB VRAM), and all deep models were trained with the same hyperparameters, where the number of input features resulted from the adaptive pooling is equal to $80$ (in Fig.~\ref{fig:gcnn}, we denote it as $k$). The patch size (PS) is kept constant, and was experimentally set to $100\times 100$ due to the available VRAM (similarly, the RF was trained sequentially for each batch, as the entire training set could not be loaded into the available memory). During the training process, we utilize early stopping, with the maximum number of epochs without improving the loss value (binary cross-entropy) over the validation set $\ValidationSet$ equal to $12$. The validation set contains $8$ random training scenes (10\% of $\TrainingSet$). The maximum number of epochs was $100$. We utilized the Adam optimizer, and the batch size and learning rate were set to $100$ and $0.001$, respectively.


To evaluate the investigated models, we employ the F-score, overall accuracy (Acc), and the Matthews correlation coefficient (MCC), with the latter metric being a reliable quality metric for imbalanced classification (it was also the metric used to rank the participants in the challenge). All metrics should be maximized, with one indicating the perfect segmentation. As the input data includes a supplementary mask that indicates which pixels should be excluded from evaluation, we calculate two versions of each metric---over the entire resulting segmentation mask (Full), and with pruning such pixels (Mask). Note that the GT is available for the full scenes, and the latter metric was used during the challenge. 

\begin{table}[ht!]
\caption{The results obtained over our test S-2 image series (we report standard deviation in the subscripts). The best result is boldfaced, and the second best is underlined.}
\label{tab:results}
\centering
\setlength{\tabcolsep}{2pt}
\scriptsize
\begin{tabular}{rcccccccccc}
\Xhline{2\arrayrulewidth}
& \multicolumn{2}{c}{\textbf{MCC}} && \multicolumn{2}{c}{\textbf{F-Score}} && \multicolumn{2}{c}{\textbf{Accuracy}}\\
\cline{2-3}\cline{5-6}\cline{8-9}
\textbf{Model$\downarrow$} & \textbf{Full} & \textbf{Mask} && \textbf{Full} & \textbf{Mask} && \textbf{Full} & \textbf{Mask}\\\hline
RF & $.606_{.222}$ & $.621_{.231}$ && $.649_{.255}$ & $.663_{.263}$ && {\ul $.873_{.071}$} & {\ul $.873_{.076}$}\\
U-Net-B & $.532_{.100}$ & $.541_{.102}$ && $.627_{.103}$ & $.638_{.103}$ && $.837_{.060}$ & $.834_{.062}$\\
U-Net-TC & {\ul $.634_{.109}$} & {\ul $.646_{.112}$} && {\ul $.721_{.096}$} & {\ul $.734_{.097}$} && $.870_{.056}$ & $.870_{.056}$\\
LSTM & $.551_{.180}$ & $.559_{.185}$ && $.628_{.200}$ & $.639_{.204}$ && $.851_{.053}$ & $.848_{.055}$\\
GCNN & $\bm{.684_{.097}}$ & $\bm{.696_{.098}}$ && $\bm{.760_{.082}}$ & $\bm{.772_{.081}}$ && $\bm{.889_{.051}}$ & $\bm{.889_{.052}}$\\
\Xhline{2\arrayrulewidth}
\end{tabular}
\end{table}

\begin{figure}[ht!]
\scriptsize
\setlength{\tabcolsep}{2pt}
\newcommand{\mywidth}{0.225}
\begin{tabular}{cc}
    \includegraphics[width=\mywidth\textwidth]{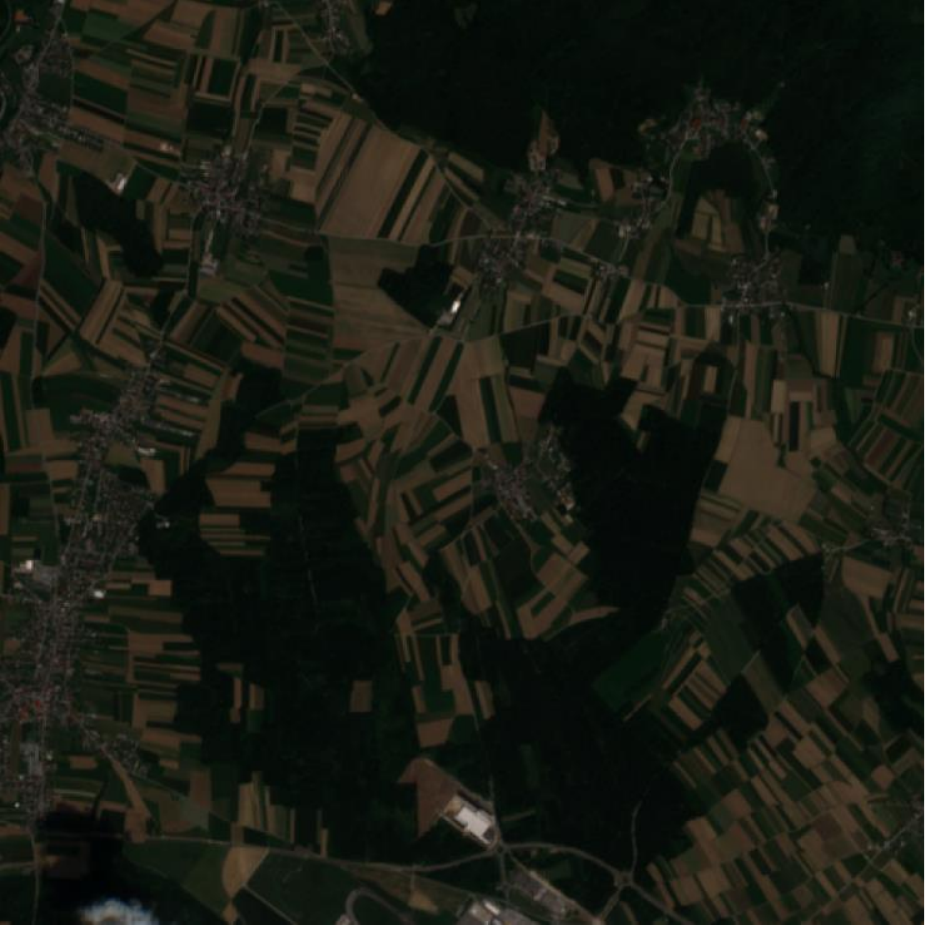} & \begin{tikzpicture}
        [,spy using outlines={circle,cyan,magnification=6,size=1.5cm, connect spies}]
        \node {\pgfimage[width=\mywidth\textwidth]{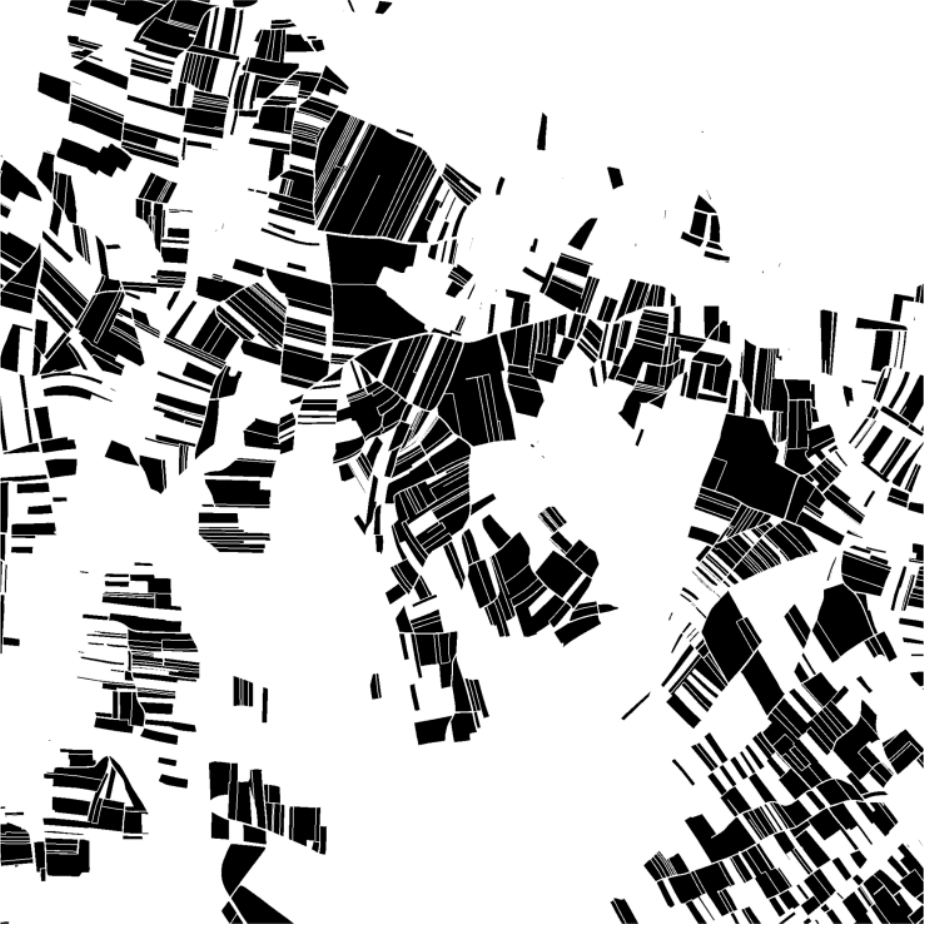}};
        \spy[every spy on node/.append style={thick},every spy in node/.append style={thick}] on (-0.7,1.2) in node [left] at (0.9,-0.7);
        \end{tikzpicture} \\
    (a) {Frame \#7} & (b) {Ground truth}\\
    \begin{tikzpicture}
        [,spy using outlines={circle,cyan,magnification=6,size=1.5cm, connect spies}]
        \node {\pgfimage[width=\mywidth\textwidth]{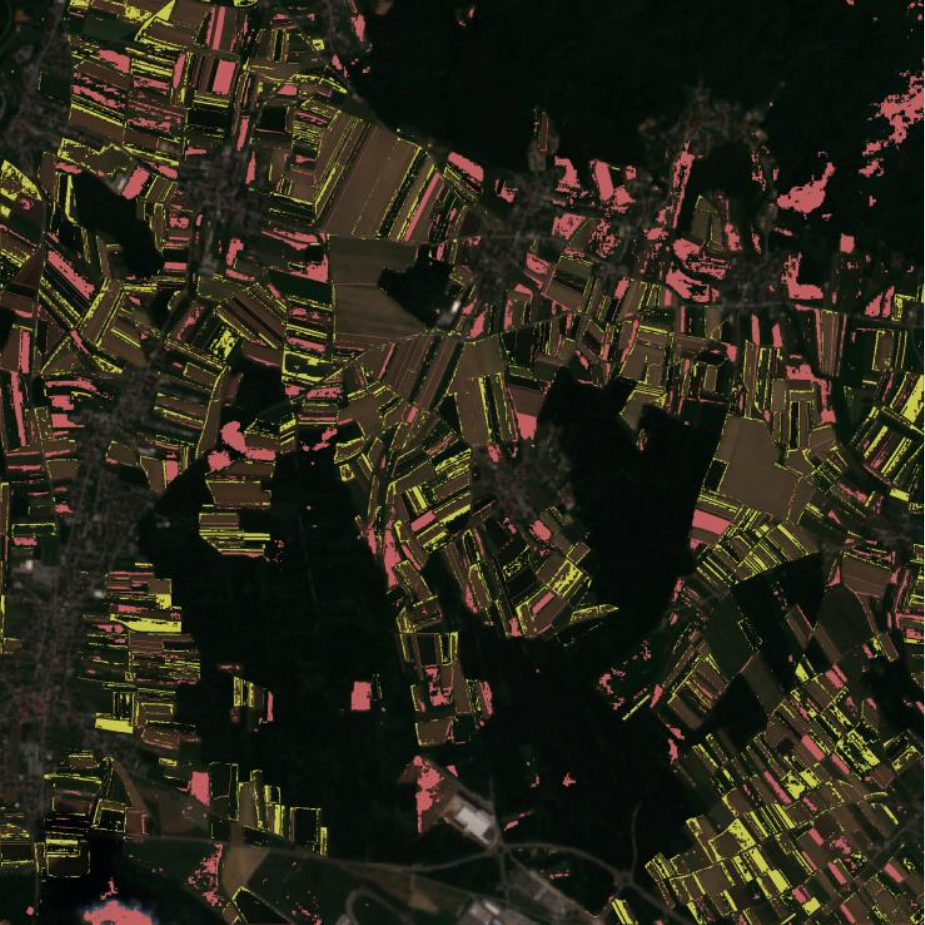}};
        \spy[every spy on node/.append style={thick},every spy in node/.append style={thick}] on (-0.7,1.2) in node [left] at (0.9,-0.7);
        \end{tikzpicture} & \begin{tikzpicture}
        [,spy using outlines={circle,cyan,magnification=6,size=1.5cm, connect spies}]
        \node {\pgfimage[width=\mywidth\textwidth]{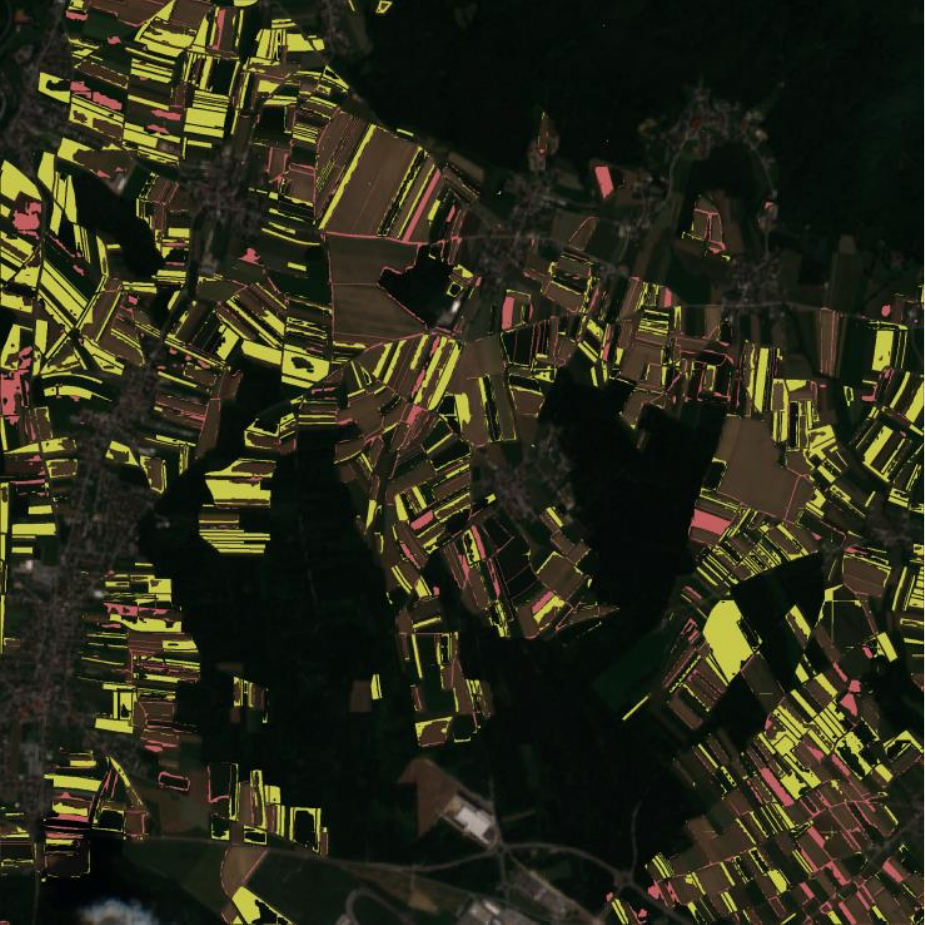}};
        \spy[every spy on node/.append style={thick},every spy in node/.append style={thick}] on (-0.7,1.2) in node [left] at (0.9,-0.7);
        \end{tikzpicture}\\
        (c) RF & (d) U-Net-TC\\
        \begin{tikzpicture}
        [,spy using outlines={circle,cyan,magnification=6,size=1.5cm, connect spies}]
        \node {\pgfimage[width=\mywidth\textwidth]{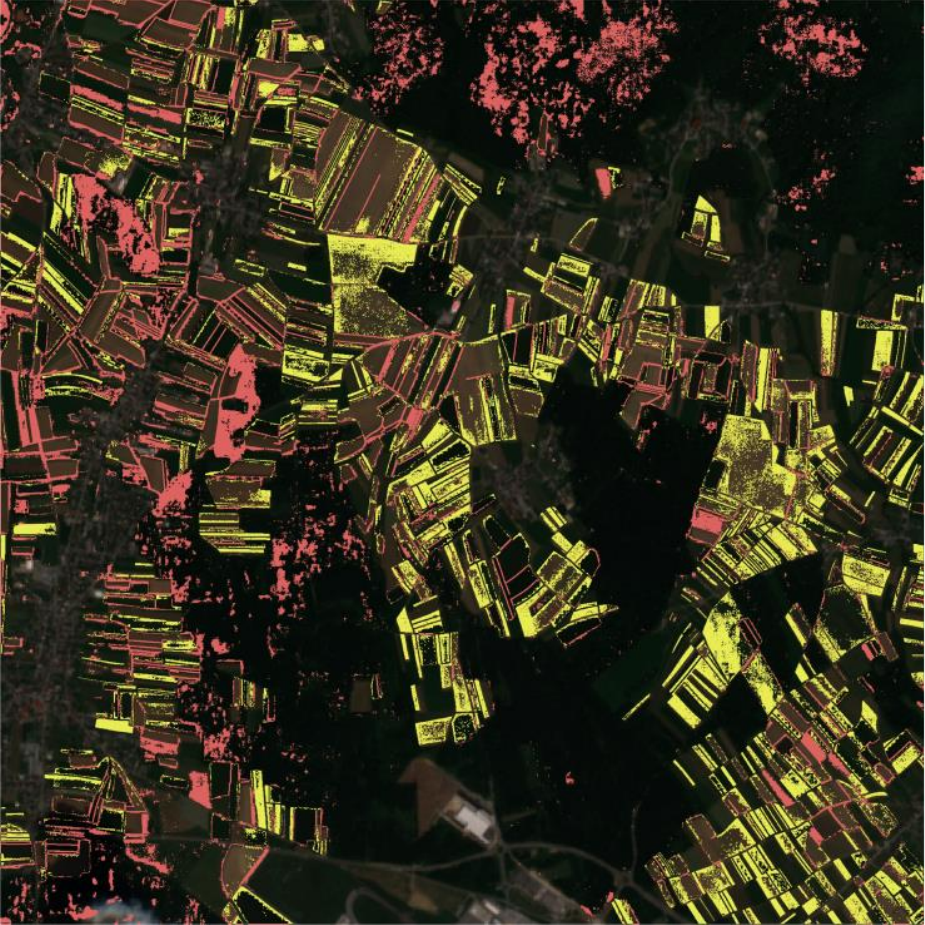}};
        \spy[every spy on node/.append style={thick},every spy in node/.append style={thick}] on (-0.7,1.2) in node [left] at (0.9,-0.7);
        \end{tikzpicture} & \begin{tikzpicture}
        [,spy using outlines={circle,cyan,magnification=6,size=1.5cm, connect spies}]
        \node {\pgfimage[width=\mywidth\textwidth]{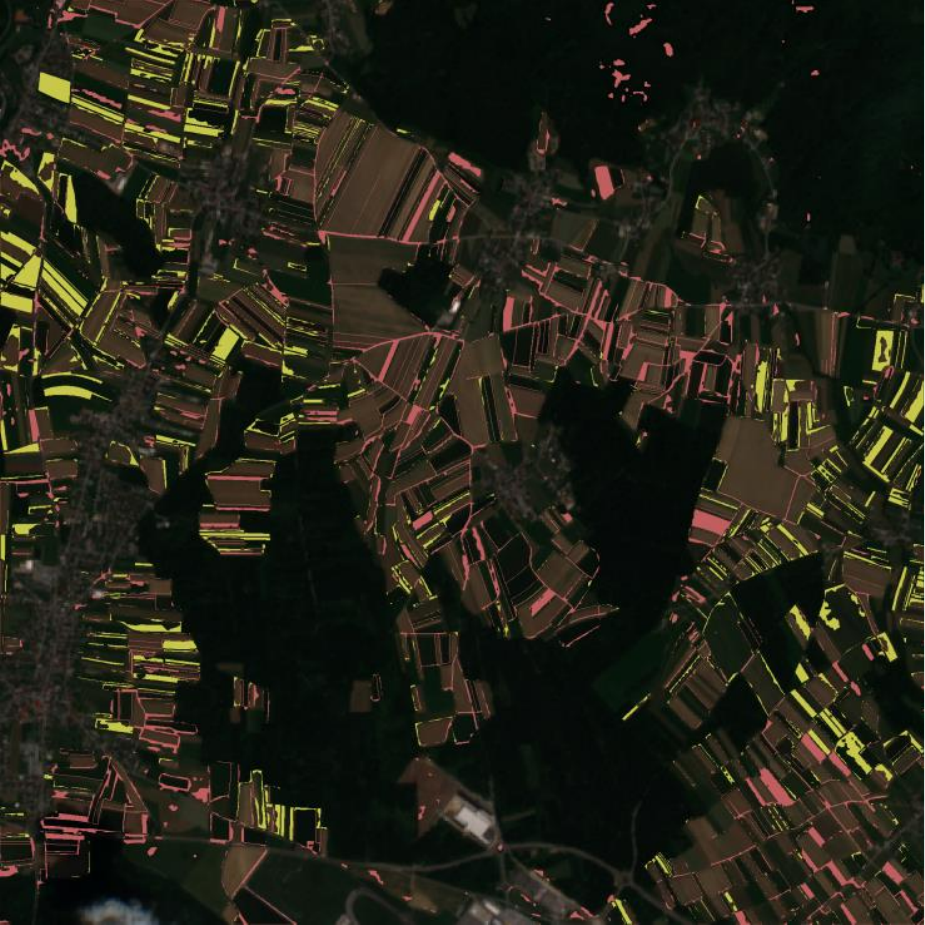}};
        \spy[every spy on node/.append style={thick},every spy in node/.append style={thick}] on (-0.7,1.2) in node [left] at (0.9,-0.7);
        \end{tikzpicture}\\
        (e) LSTM & (f) GCNN\\
\end{tabular}

    \vskip\baselineskip
    
    \centering
    \scalebox{1}{
    \scriptsize
    \setlength{\tabcolsep}{5pt}
    \begin{tabular}{rcccccccccccc} \Xhline{2\arrayrulewidth}
    & \multicolumn{2}{c}{\textbf{MCC}} && \multicolumn{2}{c}{\textbf{F-Score}} \\ 
    \cline{2-3}\cline{5-6}
    \textbf{Model$\downarrow$}    & \textbf{Full}            & \textbf{Mask}            && \textbf{Full}              & \textbf{Mask} && \textbf{Time [s]}\\ \hline
    RF & $.628$ & $.634$ && $.729$ & $.735$ && $317.1$\\
    U-Net-TC & $.640$ & $.647$ && $.717$ & $.723$ && $85.0$\\
    LSTM & $.557$ & $.565$ && $.682$ & $.690$ && $4699.4$\\
    GCNN & {$\bm{.749}$} & {$\bm{.758}$} && {$\bm{.816}$} & {$\bm{.825}$} && {$\bm{77.8}$}\\\Xhline{2\arrayrulewidth}
    \end{tabular}}
    
    \caption{Visualization of a frame (a) from an example test S-2 series (ID: \#344), the (b)~GT cultivated land segmentation with a 2.5\,m spatial resolution, and the predicted maps obtained using (c)~RF, (d)~U-Net-TC, (e)~LSTM, and (f)~the GCNN (red denotes the false positives, yellow---false negatives). We zoom a part of the image showing fine-grained details of the parcel's boundary that were correctly captured by GCNN. The best quality metrics are boldfaced.}
    \label{fig:train_with_gt}
 \end{figure}

In Table~\ref{tab:results}, we gather the results obtained for both scoring schemes (with and without masking). We can appreciate that GCNNs significantly outperformed RF, both versions of the U-Nets, and LSTMs, and delivered the highest-quality and the most stable (across the entire test set, as quantified by standard deviation) cultivated areas. The high-quality land segmentation obtained using GCNN is also manifested in Fig.~\ref{fig:train_with_gt}, where we render an example test scene segmented using all techniques (we excluded U-Net-B, as it was significantly worse than U-Net-TC), together with the GT land cultivation map (for more examples, see the supplementary material). The GCNN is able to capture subtle characteristics of the region of interest, hence precisely delineate tiny parcels (this may be pivotal in quantifying the area of such fields). GCNNs do not perform any reduction of the spatial features within the network (in contrast to U-Nets), and the edges of their binary maps appear sharper. Furthermore, the U-Net segmentation maps lack fine-grained details in those regions. It is visible in Fig.~\ref{fig:train_with_gt}(e), and this issue is resolved in GCNN, as shown in the magnified areas in Fig.~\ref{fig:train_with_gt}(f). Additionally, this visual example shows the difficulty of the task---the S-2 images within one series can vary due to changing acquisition conditions and cloud covers. Our GCNNs offer prediction faster than other methods (it is confirmed in Table~\ref{tab:parameters}; the time in Fig.~\ref{fig:train_with_gt} was measured on an Intel i7-8565U CPU, and in Table~\ref{tab:parameters}, on NVIDIA Tesla T4).

To quantify the memory requirements of the deep models in the context of deploying them in hardware-constrained execution environments, we present the number of their trainable parameters in Table~\ref{tab:parameters}. We can observe that our GCNN is a compact model, with more than $2180\times$ and $3916\times$ less parameters than U-Net-B and U-Net-TC, respectively (note that the model with the transposed convolutional layers instead of a bicubic interpolation incorporates almost two times larger number of weights). This massive difference is reflected in the size of the serialized model (33.88 kB vs. 118.65 MB, 66.09 MB and 225.46 MB for GCNN, U-Net-TC, U-Net-B and RF). It will directly impact the uplink time of such trained models to deploy them to the satellite once it is in orbit. This model update strategy\footnote{This strategy can allow us to not only replace the model with a new one, perhaps tackling a different EO task, but also to uplink an updated model for the current task (e.g.,~fine-tuned over real imagery). Thus, we aim at turning our mission into the \emph{flying laboratory}, where we can change its AI operations.} will be exploited in our Intuition-1 hyperspectral mission. Finally, we can appreciate that our GCNNs are not only compact, but also offer fast inference.
 
\begin{table}[ht!]
\centering
\caption{The number of trainable parameters in all deep models, together with the execution time averaged for the test scenes.}
\label{tab:parameters}
\scriptsize
\setlength{\tabcolsep}{10pt}
\begin{tabular}{rcc}
\Xhline{2\arrayrulewidth}
\textbf{Model} & \textbf{Trainable params} & \textbf{Inference time [s]}\\
\hline
U-Net-TC & 31,081,985 & 14.8\\
U-Net-B & 17,307,329 & 14.5\\
LSTM & 331,393 & 107.9\\
GCNN & 7,937 & 14.1\\
\Xhline{2\arrayrulewidth}
\end{tabular}
\end{table}

\begin{figure}[ht!]
\centering
\scriptsize
\begin{subfigure}[b]{\textwidth}
\begin{tikzpicture}
\begin{axis}[
    xlabel={Epochs},
    ylabel={MCC},
    ytick={0, 0.2, 0.4, 0.6, 0.8, 1.0},
    xmin=0, xmax=75,
    ymin=0, ymax=1,
    legend style={nodes={scale=0.5, transform shape}, at={(0.5,1.5)},anchor=north, font=\Large},
    legend columns=3,
    ymajorgrids=true,
    grid style=dashed,
    height=4cm,
    width=9cm
    ]
\addplot[color=blue, mark=none] coordinates {
(0.0, 0.027395860407609864) (1.0, 0.13711251909473687) (2.0, 0.3639939963937113) (3.0, 0.5640508766715407) (4.0, 0.5686986925161387) (5.0, 0.6525574872527663) (6.0, 0.6532532793611345) (7.0, 0.6845425470605063) (8.0, 0.6747581268527258) (9.0, 0.6816830331598415) (10.0, 0.6949078005083388) (11.0, 0.6723352708796295) (12.0, 0.6920053352985325) (13.0, 0.6761974835440868) (14.0, 0.696510628154954) (15.0, 0.7201205220740208) (16.0, 0.7059803306745218) (17.0, 0.7161016066597123) (18.0, 0.7372974935315474) (19.0, 0.7434728334469836) (20.0, 0.7324466983117724) (21.0, 0.7353308884157319) (22.0, 0.7465937364848612) (23.0, 0.7454509955701999) (24.0, 0.7266757016298525) (25.0, 0.7362019511878719) (26.0, 0.7278545343791208) (27.0, 0.7334889083241948) (28.0, 0.7304396699013086) (29.0, 0.7469054742698892) (30.0, 0.7241713117863489) (31.0, 0.7508406528732096) (32.0, 0.7386884001166502) (33.0, 0.7417612373534743) (34.0, 0.7581716397880952) (35.0, 0.7295543476019493) (36.0, 0.7566205556572956) (37.0, 0.7519390826790803) (38.0, 0.7659931645314063) (39.0, 0.7534623319402647) (40.0, 0.7707815344217057) (41.0, 0.7497754734702442) (42.0, 0.7756886315006649) (43.0, 0.751642325775463) (44.0, 0.7430588044792582) (45.0, 0.7539560061127145) (46.0, 0.7486239854153665) (47.0, 0.7192087273800081) (48.0, 0.7444607399222936) (49.0, 0.764296094485296) (50.0, 0.777591850607894) (51.0, 0.7440563123132886) (52.0, 0.7558592088120932) (53.0, 0.7759379083335536) (54.0, 0.7433447109729647) (55.0, 0.7610112054407528) (56.0, 0.7656892068604023) (57.0, 0.7419183692923612) (58.0, 0.7550173073183141) (59.0, 0.7498283938930613) (60.0, 0.7817675053043392) (61.0, 0.7623726323239187) (62.0, 0.7557031908580604) (63.0, 0.7392849889970599) (64.0, 0.7644763486391037) (65.0, 0.739206329054404) (66.0, 0.7724784953090239) (67.0, 0.744098828415044) (68.0, 0.7583862900196098)
    };
\addplot[color=cyan, mark=none] coordinates {
(0.0, 0.08655854115334015) (1.0, 0.38003704910995983) (2.0, 0.5724450395093343) (3.0, 0.6069740375075992) (4.0, 0.6347128637056751) (5.0, 0.6892963650276114) (6.0, 0.6678086329532258) (7.0, 0.7019837126329278) (8.0, 0.7009921368169929) (9.0, 0.6985963890669202) (10.0, 0.6972436025567859) (11.0, 0.7049195568777045) (12.0, 0.724842117736759) (13.0, 0.6574530079587722) (14.0, 0.7270647906421277) (15.0, 0.7227056895394719) (16.0, 0.7161976024547665) (17.0, 0.7240711591652902) (18.0, 0.7490604039497829) (19.0, 0.7541665766535024) (20.0, 0.7521813834751371) (21.0, 0.7477586175447373) (22.0, 0.7421178194590257) (23.0, 0.7526393113901109) (24.0, 0.7561603622831837) (25.0, 0.7392899762470803) (26.0, 0.7404267008189629) (27.0, 0.7634789297759628) (28.0, 0.7447788062060535) (29.0, 0.7517372643246814) (30.0, 0.7265767637319629) (31.0, 0.7534427248005701) (32.0, 0.7480984547818895) (33.0, 0.7506513180904402) (34.0, 0.7600676508702622) (35.0, 0.7380983505168952) (36.0, 0.7571479605025926) (37.0, 0.7615483612206095) (38.0, 0.7702168936917606) (39.0, 0.7532415314055207) (40.0, 0.7814866280736487) (41.0, 0.7608363261130812) (42.0, 0.7764246844456677) (43.0, 0.7594120729755922) (44.0, 0.746033567014538) (45.0, 0.7599572591618391) (46.0, 0.7398042536225559) (47.0, 0.7407340780541851) (48.0, 0.7399177760991416) (49.0, 0.770289891198459) (50.0, 0.7572262804310329) (51.0, 0.7517754527043923) (52.0, 0.7521659522665352) (53.0, 0.7796311847944858) (54.0, 0.7486775586063255) (55.0, 0.7772934234088044) (56.0, 0.7620649857339109) (57.0, 0.757266756432162) (58.0, 0.7448435977343217) (59.0, 0.7433008962736849) (60.0, 0.7813743327578195) (61.0, 0.7650016152504646) (62.0, 0.7661371515233665) (63.0, 0.759981073590795) (64.0, 0.7599224350254107)
    };
\addplot[color=green, mark=none] coordinates {
(0.0, 0.6090708394802382) (1.0, 0.6883062746695494) (2.0, 0.7310239002563461) (3.0, 0.7251462069616097) (4.0, 0.744094635888213) (5.0, 0.7400729173190148) (6.0, 0.7485362262092714) (7.0, 0.7656745855740686) (8.0, 0.7642810419196162) (9.0, 0.7742184814986081) (10.0, 0.7398436168830967) (11.0, 0.7321239494551529) (12.0, 0.7520868844751443) (13.0, 0.7238913549895296) (14.0, 0.7394256327945589) (15.0, 0.7546344921566897) (16.0, 0.7399571230337456) (17.0, 0.7504970657868896) (18.0, 0.7674705422344126) (19.0, 0.7580042302597259) (20.0, 0.7548817868761966) (21.0, 0.7834672150621791)
    };
\addplot[color=orange, mark=none] coordinates {
(0.0, 0.5949380384358582) (1.0, 0.6927429192963437) (2.0, 0.67739396122325) (3.0, 0.6996592744904226) (4.0, 0.7209154879477377) (5.0, 0.7073392710196632) (6.0, 0.71399475912478) (7.0, 0.7419350291644539) (8.0, 0.7581037415206829) (9.0, 0.7473628495552903) (10.0, 0.7177407013317648) (11.0, 0.7341549915881982) (12.0, 0.737508123343614) (13.0, 0.7045045062625326) (14.0, 0.7225115119858846) (15.0, 0.7160279922581009) (16.0, 
0.747009247886369) (17.0, 0.7493542292712456) (18.0, 0.7570047657996641) (19.0, 0.719062772399981) (20.0, 0.732131341438187) (21.0, 0.7473055029456706) (22.0, 0.7619591499644589) (23.0, 0.7688017020496675) (24.0, 0.757596090221079) (25.0, 0.7457728880500782) (26.0, 0.7492000038731575) (27.0, 0.761113574338506) (28.0, 0.7435393453315027) (29.0, 0.7513381464750244) (30.0, 0.7250526624488148) (31.0, 0.7425360571512347) (32.0, 0.7479959111514246) (33.0, 0.7311957211757902) (34.0, 0.7567889263578143) (35.0, 0.7709684259363792) (36.0, 0.7482138785491728) (37.0, 0.7276238941294126) (38.0, 0.7635121582944082) (39.0, 0.7734931514805183) 
(40.0, 0.7797947464726315) (41.0, 0.7461862628507892) (42.0, 0.7697849047414487) (43.0, 0.7470030600848829) (44.0, 0.721110676096151) (45.0, 0.7431367463378511) (46.0, 0.7541388011590335) (47.0, 0.7395793164157667) (48.0, 0.7217810185166351) (49.0, 0.7573984825264586) (50.0, 0.767908607243657) (51.0, 0.7335479228814665) (52.0, 0.7513871950513165)
    };
\addplot[color=magenta, mark=none] coordinates {
(0.0, 0.5416126734751419) (1.0, 0.6841925158810716) (2.0, 0.6851513885838978) (3.0, 0.6925100383609337) (4.0, 0.6617817070084587) (5.0, 0.7189789166175089) (6.0, 0.6441917223707869) (7.0, 0.7212940486533055) (8.0, 0.7333532906934839) (9.0, 0.7188437461475214) (10.0, 0.673390010906378) (11.0, 0.7026175839566713) (12.0, 0.7206886111252686) (13.0, 0.6932146030254873) (14.0, 0.6956763125906431) (15.0, 0.7280488594882535) (16.0, 0.7325665249599048) (17.0, 0.721318166156824) (18.0, 0.7456193580971301) (19.0, 0.7181700269328092) (20.0, 0.7182540610740483) (21.0, 0.7440819132995455) (22.0, 0.7442638220854753) (23.0, 0.7674388779987338) 
(24.0, 0.7517634820507768) (25.0, 0.7286540047960441) (26.0, 0.7476545108990451) (27.0, 0.7455887132745629) (28.0, 0.7173235018586552) (29.0, 0.716639505399647) (30.0, 0.7321762658444221) (31.0, 0.7270456612032334) (32.0, 0.7425001612734664) (33.0, 0.7239480896493762) (34.0, 0.755264013290484) (35.0, 0.7283658116880867) (36.0, 0.7599901547645808) (37.0, 0.707773247684029) (38.0, 0.7589164116165028) (39.0, 0.7705403024918095) (40.0, 0.7779397725420663) (41.0, 0.723188390595311) (42.0, 0.7569568822323631) (43.0, 0.7087142527783185) (44.0, 0.7130315124507156) (45.0, 0.7322991116943206) (46.0, 0.7578823780069583) (47.0, 0.728202329519653) (48.0, 0.7167758852994471) (49.0, 0.7440069496442006) (50.0, 0.7638573225973173) (51.0, 0.7270549103772705) (52.0, 0.733850363189995) (53.0, 0.7461873982222442) (54.0, 0.7356536141579405) (55.0, 0.7149373021485756) (56.0, 0.7125598920600629) (57.0, 0.7325008859693153) (58.0, 0.7528608509866673) (59.0, 0.7445748429133775) (60.0, 0.7672022529191447) (61.0, 0.7223292044433284) (62.0, 0.76067272089784) (63.0, 0.7379642196064806) (64.0, 0.7413071433017733) (65.0, 0.7509716904453057) (66.0, 0.7576391722549872) (67.0, 0.7520548461706296) (68.0, 0.7587138142210557) (69.0, 0.74592354032626) (70.0, 0.7712198207416926) (71.0, 0.7776944878169024) (72.0, 0.7533444510260958) (73.0, 0.7489431836360129) (74.0, 0.7086268178368602)
    };
\addplot[color=black, mark=none] coordinates {
(0.0, 0.44263206706652974) (1.0, 0.46718188092012686) (2.0, 0.48465376899827295) (3.0, 0.5116810575996046) (4.0, 0.5446596932714766) (5.0, 0.616161253975686) (6.0, 0.3972078847328274) (7.0, 0.587318606589802) (8.0, 0.4566270322466389) (9.0, 0.5773132181534757) (10.0, 0.539479347317557) (11.0, 0.38934892422295464) (12.0, 0.48831721374517933) (13.0, 0.628720863924243) (14.0, 0.4999236633627142) (15.0, 0.640220995841691) (16.0, 0.6496776104771428) (17.0, 0.6184239230835551) (18.0, 0.47109098443653175) (19.0, 0.6239016533756883) (20.0, 0.5902041095341669) (21.0, 0.5496558214174547) (22.0, 0.6106750848041615) (23.0, 0.5234810752057792) (24.0, 0.5824005844436966) (25.0, 0.6146058340018634) (26.0, 0.6036345856523591) (27.0, 0.5237660058290047) (28.0, 0.5870300647341198) (29.0, 0.5960489956921908) (30.0, 0.6796666718237117) (31.0, 0.6314515154969831) (32.0, 0.6196992146903504) (33.0, 0.6324171212016803) (34.0, 0.6970064829322065) (35.0, 0.6394635532324735) (36.0, 0.5814595092645449) (37.0, 0.6717336683174672) (38.0, 0.5514459997523654) (39.0, 0.6786577712117488) (40.0, 0.664601585862503) (41.0, 0.7092089346643629) (42.0, 0.5678330260993497) (43.0, 0.6776344490418298) (44.0, 0.6478777385494863) (45.0, 0.6732980083432016) (46.0, 0.60622494895306)
    };
\addplot[color=gray, mark=none] coordinates {
(0.0, 0.5039695998670048) (1.0, 0.5894734725875489) (2.0, 0.5366340685064854) (3.0, 0.6519916907900816) (4.0, 0.5527226473639155) (5.0, 0.5766285516671597) (6.0, 0.4964710948592498) (7.0, 0.5401888806453761) (8.0, 0.5984643950881767) (9.0, 0.5614259643857074) (10.0, 0.5741468252266967) (11.0, 0.5928618007710622) (12.0, 0.5687911407772134) (13.0, 0.6326807854349004) (14.0, 0.5744544256142893) (15.0, 0.6148534733673165) (16.0, 0.5399290583906444) (17.0, 0.5896013945461651) (18.0, 0.5844402395420025) (19.0, 0.6062548973123465) (20.0, 0.5749781618852646) (21.0, 0.5766854220503439) (22.0, 0.6207275616265165) (23.0, 0.6023230339734125) (24.0, 0.6391004827519655) (25.0, 0.4333736100183004) (26.0, 0.5998440193480864) (27.0, 0.6047583151419351)
    };
\addplot[color=brown, mark=none] coordinates {
    (0.0, 0.14994592359308045) (1.0, 0.18071942726599888) (2.0, 0.16187361406525702) (3.0, 0.35962602448721426) (4.0, 0.3337282572858687) (5.0, 0.28840453376131825) (6.0, 0.4513136399930141) (7.0, 0.38041115089205846) (8.0, 0.3995929495404029) (9.0, 0.41736979713491107) (10.0, 0.381233662901303) (11.0, 0.4053491495966265) (12.0, 0.45867746942061266) (13.0, 0.4288669155720536) (14.0, 0.4587266329489763) (15.0, 0.39403130123108715) (16.0, 0.447899524645787) (17.0, 0.42873944928058344)
};
\legend{PS: 100, PS: 80, PS: 40, PS: 20, PS: 10, U-Net-TC, U-Net-B, LSTM}
\end{axis}
\end{tikzpicture}
\end{subfigure}
\centering
\begin{subfigure}[b]{\textwidth}
\begin{tikzpicture}
\begin{axis}[
    xlabel={Epochs},
    ylabel={BCE},
    ytick={0, 0.2, 0.4, 0.6, 0.8, 1.0},
    xmin=0, xmax=75,
    ymin=0, ymax=1,
    ymajorgrids=true,
    grid style=dashed,
    height=4cm,
    width=9cm
    ]
\addplot[color=blue, mark=none] coordinates {
(0.0, 0.5431148260831833) (1.0, 0.4959254441782832) (2.0, 0.49133785907179117) (3.0, 0.39462773501873016) (4.0, 0.3765806835144758) (5.0, 0.3094360902905464) (6.0, 0.29980515129864216) (7.0, 0.3099737814627588) (8.0, 0.3040202911943197) (9.0, 0.28797291219234467) (10.0, 0.29765187483280897) (11.0, 0.2949035745114088) (12.0, 0.29095616191625595) (13.0,0.28456185944378376) (14.0, 0.27933910535648465) (15.0,0.27690936205908656) (16.0, 0.2737453072331846) (17.0, 0.27816205378621817) (18.0, 0.2637609438970685) (19.0, 0.25611676182597876) (20.0, 0.26321439584717155) (21.0, 0.2574770199134946) (22.0, 0.24979799566790462) (23.0,0.26841727970167994) (24.0, 0.26502613676711917) (25.0,0.24453243473544717) (26.0, 0.25050853146240115) (27.0, 0.2606280446052551) (28.0, 0.2618295238353312) (29.0, 0.25613387720659375) (30.0, 0.2447687853127718) (31.0, 0.25214861892163754) (32.0, 0.23997073993086815) (33.0,0.23740956094115973) (34.0, 0.23405370861291885) (35.0,0.25655221845954657) (36.0, 0.24175839452072978) (37.0,0.24440842028707266) (38.0, 0.2337902751751244) (39.0, 0.24814758822321892) (40.0, 0.23138112016022205) (41.0,0.23526368895545602) (42.0, 0.23645998444408178) (43.0, 0.2386440229602158) (44.0, 0.2317761075682938) (45.0, 0.23746443074196577) (46.0, 0.24858620576560497) (47.0,0.26075572380796075) (48.0, 0.24282869789749384) (49.0,0.24961534002795815) (50.0, 0.22485336381942034) (51.0,0.24816730013117194) (52.0, 0.2263358396012336) (53.0, 0.22952593630179763) (54.0, 0.23529249196872115) (55.0,0.23316501500084996) (56.0, 0.22333113476634026) (57.0,0.23848305642604828) (58.0, 0.23764612013474107) (59.0, 0.2399330553598702) (60.0, 0.2340122601017356) (61.0, 0.22868111380375922) (62.0, 0.2299410467967391) (63.0, 0.2478275429457426) (64.0, 0.23376885848119855) (65.0, 0.2408427339978516) (66.0, 0.23154430370777845) (67.0,0.23636808479204774) (68.0, 0.23331185104325414)
    };
\addplot[color=cyan, mark=none] coordinates {
(0.0, 0.5622334486016861) (1.0, 0.4515156854803746) (2.0, 0.34725415821258837) (3.0, 0.3705897915821809) (4.0, 0.36064440585099733) (5.0, 0.29281649423333317) (6.0, 0.2954090234751885) (7.0, 0.2868603551043914) (8.0, 0.29147859347554356) (9.0, 0.2714430414713346) (10.0, 0.28528332309081006) (11.0, 0.2588469828837193) (12.0, 0.2542874331657703) (13.0, 0.2949515523818823) (14.0, 0.2555985843332914) (15.0, 0.26413980241005236) (16.0, 0.27018480891218555) (17.0,0.25582094289935553) (18.0, 0.25108394485253555) (19.0,0.24944705688036406) (20.0, 0.245178516381062) (21.0, 0.2479793162873158) (22.0, 0.2619003805403526) (23.0, 0.2529569285420271) (24.0, 0.24234921542497778) (25.0, 0.2445718886760565) (26.0, 0.240338953355184) (27.0, 0.24527704944977394) (28.0, 0.25073123895204985) (29.0,0.25988514664081425) (30.0, 0.23911319415156657) (31.0, 0.2491854936457597) (32.0, 0.24086215232427302) (33.0,0.23380480534755266) (34.0, 0.2384014459183583) (35.0, 0.24762317624229652) (36.0, 0.2413667916105344) (37.0, 0.22889399901032448) (38.0, 0.23904141396857226) (39.0, 0.2509872377491914) (40.0, 0.22710491645221528) (41.0,0.23793431414434543) (42.0, 0.2286726783674497) (43.0, 0.2382735048349087) (44.0, 0.23399172064203483) (45.0,0.24054720195440146) (46.0, 0.2568818161693903) (47.0, 0.2528795769008306) (48.0, 0.24339411407709122) (49.0,0.23273554673561683) (50.0, 0.23838214518932196) (51.0,0.24893367147216433) (52.0, 0.22587872210603493) (53.0,0.22871883356800446) (54.0, 0.2313475153194024) (55.0, 0.2277944947664554) (56.0, 0.22869243358190244) (57.0,0.23167489397411162) (58.0, 0.24154509317416412) (59.0,0.24211728314940745) (60.0,0.23587136472073886) (61.0, 0.232022057645596) (62.0, 0.22690086238659346) (63.0,0.22739828635866824) (64.0, 0.2375268661058866)
    };
\addplot[color=green, mark=none] coordinates {
(0.0, 0.34625504925847056) (1.0, 0.28518720902502537) (2.0, 0.26071246959269045) (3.0, 0.2629935158789158) (4.0, 0.26371943645179274) (5.0, 0.2540398545563221) (6.0, 0.23977505281567574) (7.0, 0.24564668636769058) (8.0, 0.24051574036478998) (9.0, 0.22698630336672068) (10.0, 0.2535185355693102) (11.0, 0.25572473343461755) (12.0, 0.24150818809866903) (13.0,0.24846582166850567) (14.0, 0.24250496946275235) (15.0,0.24132906913757324) (16.0, 0.2506868913769722) (17.0, 0.24527016103267668) (18.0, 0.24488061241805553) (19.0,0.24619366366416215) (20.0, 0.2463974764570594) (21.0, 0.2296568978577852)
    };
\addplot[color=orange, mark=none] coordinates {
(0.0, 0.348683987557888) (1.0, 0.27930742748081683) (2.0, 0.28470359230414033) (3.0, 0.270597905870527) (4.0, 0.25815369734540583) (5.0, 0.26858384687453507) (6.0, 0.24759631961584092) (7.0, 0.24540440157055854) 
(8.0, 0.23369681572541595) (9.0, 0.23380220390856266) (10.0, 0.2613628001883626) (11.0, 0.23747379631735382) (12.0, 0.23332464121282104) (13.0, 0.25508287088014187) (14.0, 0.2500299080926925) (15.0, 0.2511536198388785) (16.0, 0.23029724948108193) (17.0, 0.23045691381208597) (18.0, 0.2352706207148731) (19.0, 0.27059081396088003) (20.0, 0.2479067488759756) (21.0, 0.23725465564988552) (22.0, 0.23258608386851848) (23.0, 0.22514720724895598) (24.0, 0.22707348322495818) (25.0, 0.23122343349270522) (26.0, 0.22370862907730044) (27.0, 0.225262269563973) (28.0, 0.23951155746355654) (29.0, 0.24194603213109078) (30.0, 0.2249482957366854) (31.0, 0.24746442524716258) (32.0, 0.2195468735136092) (33.0, 0.2405985471745953) (34.0, 0.22889943329617382) (35.0, 0.22612869078293443) (36.0, 0.22804740409366786) (37.0, 0.25824101417325435) (38.0, 0.22824763891287148) (39.0, 0.21903644176200032) (40.0, 0.21463148430921136) (41.0, 0.22947037108242513) (42.0, 0.22526865119114517) (43.0, 0.2390194412693381) (44.0, 0.22893911423161625) (45.0, 0.2471987154800445) (46.0, 0.21943743257783355) (47.0, 0.2422416797466576) (48.0, 0.240210465118289) (49.0, 0.24081907040439546) (50.0, 0.21810851029120387) (51.0, 0.2425080991256982) (52.0, 0.21967838148586452)    
    };
\addplot[color=magenta, mark=none] coordinates {
(0.0, 0.42448071468621484) (1.0, 0.27320782621391115) (2.0, 0.28503550452180204) (3.0, 0.29730142882093785) (4.0, 0.31054462682455775) (5.0, 0.28712843771791086) (6.0, 0.34382798923878) (7.0, 0.2805918271513656) 
(8.0, 0.24545193039113655) (9.0, 0.2625707303499803) (10.0, 0.30323298868723214) (11.0, 0.278998942009639) (12.0, 0.2596767954947427) (13.0, 0.2822915086394641) (14.0, 0.2751600187446457) (15.0, 0.25565146846929565) (16.0, 0.2424417601292953) (17.0, 0.2612770895054564) (18.0, 0.2546408556564711) (19.0, 0.2657232121401466) (20.0, 0.25789540586061777) (21.0, 0.2458270757668652) (22.0, 0.24844641459174455) (23.0, 0.2295386323519051) (24.0, 0.22773078374797479) (25.0, 0.23492561067920179) (26.0, 0.23023540294263511) (27.0, 0.2429971246561035) (28.0, 0.2547514452622272) (29.0, 0.2865981615916826) (30.0, 0.2291203343286179) (31.0, 0.265034638926154) (32.0, 0.22042785195517356) (33.0, 0.24743716478231362) (34.0, 0.2413432088168338) (35.0, 0.2758658623951487) (36.0, 0.2193057579197921) (37.0, 0.26657352678012103) (38.0, 0.24789441435365006) (39.0, 0.22001818934106265) (40.0, 0.2189090675767511) (41.0, 0.2435326590575278) (42.0, 0.23063882965128868) (43.0, 0.2564084546943195) (44.0, 0.22961914866464211) (45.0, 0.2475790275633335) (46.0, 0.21954386867582795) (47.0, 0.24259724187664689) (48.0, 0.2550570281525143) (49.0, 0.24744921536883335) (50.0, 0.21839816656196487) (51.0, 0.23816790555487388) (52.0, 0.260376629132079) (53.0, 0.22916873807436788) (54.0, 0.23652375509147533) (55.0, 0.251317134660203) (56.0, 0.2502478082408197) (57.0, 0.2319712109141983) (58.0, 0.22254803852643817) (59.0, 0.23005667706485838) (60.0, 0.2300451277755201) (61.0, 0.2477585218544118) (62.0, 0.21490230772993527) (63.0, 0.2379428638902027) (64.0, 0.243736264440231) (65.0, 0.2274688621610403) (66.0, 0.23667024986818433) (67.0, 0.23169549211976118) (68.0, 0.22667884515365588) (69.0, 0.22163738815812392) 
(70.0, 0.220817986180773) (71.0, 0.22204982792260125) (72.0, 0.2268837195634842) (73.0, 0.2325634292070754) (74.0, 0.2611358099873178)
    };
    \addplot[color=black, mark=none] coordinates {
(0.0, 0.4265193408355117) (1.0, 0.4502489739097655) (2.0, 0.49723039753735065) (3.0, 0.4769162442535162) (4.0, 0.3756165229715407) (5.0, 0.3592635029926896) (6.0, 0.4436869490891695) (7.0, 0.37066698633134365) (8.0, 0.4396074693650007) (9.0, 0.3739985032007098) (10.0, 0.3684781966730952) (11.0, 0.6349701276049018) (12.0, 0.40073699736967683) (13.0, 0.3572740089148283) (14.0, 0.4037946406751871) (15.0, 0.3154674982652068) (16.0, 0.30632736161351204) (17.0, 0.31064418237656355) (18.0, 0.6110233496874571) (19.0, 0.333038330078125) (20.0, 0.32312041660770774) (21.0, 0.3380983080714941) (22.0, 0.3042493653483689) (23.0, 0.3633154481649399) (24.0, 0.3346468494273722) (25.0, 0.33364866580814123) (26.0, 0.3396229459904134) (27.0, 0.3765082210302353) (28.0, 0.34621481923386455) (29.0, 0.386203121393919) (30.0, 0.27792519237846136) (31.0, 0.3030525869689882) (32.0, 0.343841350171715) (33.0, 0.306527650449425) (34.0, 0.26813531713560224) (35.0, 0.28819648222997785) (36.0, 0.370487904176116) (37.0, 0.2985438872128725) (38.0, 0.6369596268050373) (39.0, 0.27143466053530574) (40.0, 0.2809435874223709) (41.0, 0.2752090450376272) (42.0, 0.3865173449739814) (43.0, 0.2765076383948326) (44.0, 0.2841232046484947) (45.0, 0.292067002505064) (46.0, 0.4099174099974334)
    };
    \addplot[color=gray, mark=none] coordinates {
(0.0, 0.49365153163671494) (1.0, 0.34033961687237024) (2.0, 0.4118089256808162) (3.0, 0.3229460557922721) (4.0, 0.43043763749301434) (5.0, 0.3718542670831084) (6.0, 0.4465863509103656) (7.0, 0.3801419483497739) (8.0, 0.3688534954562783) (9.0, 0.3782347096130252) (10.0, 0.362180357798934) (11.0, 0.3447964326478541) (12.0, 0.4935250850394368) (13.0, 0.3252723994664848) (14.0, 0.3475668132305145) (15.0, 0.3211395009420812) 
(16.0, 0.39966526813805103) (17.0, 0.3826673999428749) (18.0, 0.3560009473003447) (19.0, 0.3394591845571995) (20.0, 0.37951321993023157) (21.0, 0.3439554809592664) (22.0, 0.3278301148675382) (23.0, 0.3589379917830229) (24.0, 0.32292181393131614) (25.0, 0.4397294241935015) (26.0, 0.3891742220148444) (27.0, 0.3617364796809852)
    };
\addplot[color=brown, mark=none] coordinates {
    (0.0, 0.5305924145528115) (1.0, 0.4536714883637614) (2.0, 0.4840242457564454) (3.0, 0.4599701685388572) (4.0, 0.4034269300615415) (5.0, 0.3751176951182424) (6.0, 0.38327174252131957) (7.0, 0.3836612026253715) (8.0, 0.3583389240916585) (9.0, 0.35413296524347965) (10.0, 0.36648404786537864) (11.0, 0.3684009374288143) (12.0, 0.34346914057416145) (13.0, 0.35728311376413335) (14.0, 0.335056409044555) (15.0, 0.3801106883765169) (16.0, 0.35193038010402233) (17.0, 0.3596615210513119)
};
\end{axis}
\end{tikzpicture}
\end{subfigure}
\caption{The MCC and binary cross-entropy (BCE) scores obtained over the validation set $\ValidationSet$ by the GCNNs trained with different patch sizes (PS), and by both U-Nets. The varying length of the training curves indicates that the early stopping condition has been met.}
\label{fig:val_curves}
\end{figure}
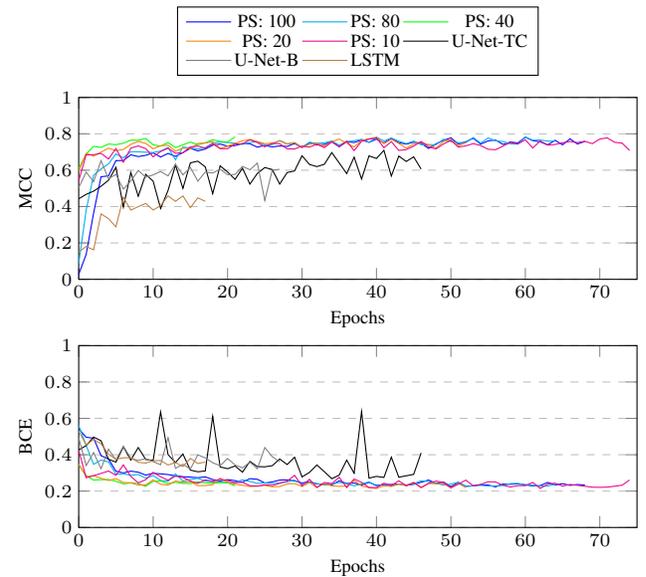

To verify the impact of the patch size on the GCNN training, we show MCC and binary cross-entropy obtained for $\ValidationSet$ for all epochs in Fig.~\ref{fig:val_curves}. PS affects the number of input nodes of GCNN, and---when coupled with the constant batch size---directly influences training. We can appreciate that the final MCC and cross-entropy are high for small PS values too, thus our GCNN is robust against a range of values of this hyperparameter and always delivers accurate segmentation. On the other hand, MCC and cross-entropy for both U-Nets indicate that these networks converge faster to the worse models. In the supplement, we can see that changing the bicubic interpolation to other techniques (such as nearest-neighbor or Lanczos) has the minimal impact on the training.

Finally, we utilized all models to segment the test data delivered by the challenge organizers---the weighted MCC scores\footnote{The organizers of the challenge report the modified MCC metric which assigns a higher weight to small parcels.} calculated by the validation server amounted to 0.597, 0.475, 0.544, 0.543, and 0.635 for RF, U-Net-B, U-Net-TC, LSTM and GCNN, respectively (for GCNN with the nearest-neighbor, Lanchos and inter-area interpolations we got 0.623, 0.610, and 0.611). To further improve these results, it would be pivotal to increase the patch size fed to the models (due to the VRAM limitations, we trained and infer over $100\times 100$ patches, as mentioned earlier). For the U-Nets (with 722,615 trainable parameters, roughly $91\times$ more than GCNN) trained over the full scenes ($500\times 500$ pixels) using the GPU server provided by the organizers, we obtained MCC of approximately 0.663. Furthermore, we were able to train a more complex version of U-Net (30,601,221 parameters, $3856\times$ more than GCNN), incorporating an already upscaled image of size $2000\times 2000$ (we used bicubic interpolation), for which MCC amounted to 0.773. These experiments show that increasing the spatial context could improve the segmentation scores---it constitutes our current efforts for GCNNs~\cite{DBLP:conf/rtas/Zhou0GQQWCDZH21}.

\section{Conclusions and Future Work}
\label{sec:conclusions}
We proposed an end-to-end processing pipeline built upon a graph convolutional network to extract precise high-resolution cultivated land segmentation maps from S-2 image series of lower spatial resolution. The experimental results, obtained over a range of scenes containing a varying number of MSIs captured in different time points revealed that our technique significantly outperforms both classical and deep learning models through delivering higher-quality segmentation maps. Additionally, we massively reduced the number of trainable parameters of the deep learning models when compared to the U-Nets (up to more than $3900\times$). This, in turn, allowed us to dramatically decrease the memory footprint of the segmentation model. Elaborating such resource-frugal deep networks is pivotal in the context of satellite missions which can benefit from uplinking a trained (or fine-tuned) model to the satellite while it is in in-orbit operation, as it directly influences the upload time. Finally, we showed that our GCNN is robust against the image patch size, and lead to precise segmentation for a range of vastly different patch sizes, and it offers fast inference. We believe that the research reported in this letter can become an important step toward deploying compact yet efficient and robust deep models in EO satellites. 

We are currently working on using the weighted edges (and heterogeneous graphs) to further exploit the temporal information, and on porting the models to Intuition-1. We focus on not only benchmarking such algorithms~\cite{rs13193981}, but also on verifying their robustness against on-board conditions~\cite{rs13081532}. Finally, we work on utilizing other interpolation techniques and our multi-image SR algorithms, and to make them task-driven for precise segmentation of satellite images.

\ifCLASSOPTIONcaptionsoff
  \newpage
\fi

\bibliographystyle{IEEEtran}
\bibliography{ref_all}

\end{document}



\begin{center} \textbf{Graph Neural Networks Extract High-Resolution Cultivated Land Maps from Sentinel-2 Image Series (Supplementary Material)}\end{center}
\begin{center} \vspace*{-0.3cm} Lukasz Tulczyjew, Michal Kawulok, Nicolas Long{\'e}p{\'e}, Bertrand Le Saux, Jakub Nalepa\end{center}
\begin{center} \vspace*{-0.5cm} \texttt{jnalepa@ieee.org}\end{center}

This supplementary material collects the training curves obtained for the proposed graph neural network models coupled with a range of interpolation techniques, together with the example Sentinel-2 scenes segmented using the proposed technique (Section~\ref{sec:exp}).

\section{Detailed Experimental Results}\label{sec:exp}

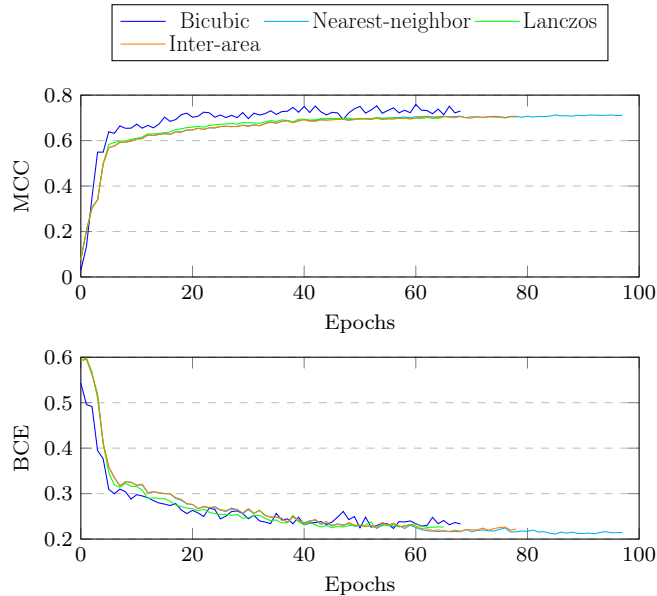
\begin{figure}[ht!]
\centering
\scriptsize
\begin{subfigure}[b]{\textwidth}
\centering
\begin{tikzpicture}
\begin{axis}[
    xlabel={Epochs},
    ylabel={MCC},
    ytick={0, 0.2, 0.4, 0.6, 0.8},
    xmin=0, xmax=100,
    ymin=0, ymax=0.8,
    legend style={nodes={scale=0.5, transform shape}, at={(0.5,1.5)},anchor=north, font=\Large},
    legend columns=3,
    ymajorgrids=true,
    grid style=dashed,
    height=4cm,
    width=9cm
    ]
\addplot[color=blue, mark=none] coordinates {
(0, 0.029484025762162976) (1, 0.1337088501562489) (2, 0.3506808369430439) (3, 0.5493088447936416) (4, 0.5490650541621941) (5, 0.6387657640874598) (6, 0.6320234875021541) (7, 0.6646568487108123) (8, 0.6540547395015528) (9, 0.654711124841611) (10, 0.6722819698857404) (11, 0.6545402026323266) (12, 0.6670717334609639) (13, 0.6559262390848895) (14, 0.6727848894334091) (15, 0.7032545210527491) (16, 0.6845327080633765) (17, 0.6924956989744419) (18, 0.7141867313885535) (19, 0.7204483948759042) (20, 0.7027664467322923) (21, 0.7071879959954293) (22, 0.7249602346845442) (23, 0.7223553312303089) (24, 0.7027928045066295) (25, 0.7126928018239563) (26, 0.7012205658193662) (27, 0.7112547646910113) (28, 0.7008367506951334) (29, 0.7232511365157384) (30, 0.6966604805117217) (31, 0.7206369883816761) (32, 0.7127447153082049) (33, 0.7181077628537715) (34, 0.7312596189553275) (35, 0.7149978712909061) (36, 0.7274829036564185) (37, 0.7291746790263157) (38, 0.7459365279954362) (39, 0.7246832048314457) (40, 0.7500572223911945) (41, 0.72356297743164) (42, 0.7519765315878503) (43, 0.7264518917288024) (44, 0.7129905489387918) (45, 0.7243255805070428) (46, 0.7215438524892318) (47, 0.692854509467241) (48, 0.7221978978224074) (49, 0.7396103315251379) (50, 0.750678310865557) (51, 0.7208822449608565) (52, 0.7346489247880028) (53, 0.7529264860729042) (54, 0.7200644695685607) (55, 0.7325273455306365) (56, 0.7438846582591039) (57, 0.7222351429077375) (58, 0.733333185376886) (59, 0.7228442007306244) (60, 0.7590291133313963) (61, 0.7339462766357493) (62, 0.7320787383162918) (63, 0.7180611886021547) (64, 0.738468320885302) (65, 0.7138116861513173) (66, 0.7511577471082939) (67, 0.7219353153227208) (68, 0.7300977016571875)
    };
\addplot[color=cyan, mark=none] coordinates {
(0.0, 0.07508574405738883) (1.0, 0.20213507784378176) (2.0, 
0.30614776031722396) (3.0, 0.3384592118527896) (4.0, 0.4971843584997781) (5.0, 0.56952801525755) (6.0, 0.5755349619369484) (7.0, 0.5922218805249062) (8.0, 0.5927222974317876) (9.0, 0.5978645267129666) (10.0, 0.6077176490586206) (11.0, 0.609376916470295) (12.0, 0.6244953401439919) (13.0, 0.6227135343824762) (14.0, 0.6265642125701483) (15.0, 0.6304503890908405) (16.0, 0.6280944506332181) (17.0, 0.6388669063590378) (18.0, 0.638355956832302) (19.0, 0.646813345982206) (20.0, 0.6469581295437716) (21.0, 0.6543986361060722) (22.0, 0.6488023283746313) 
(23.0, 0.6566181372385658) (24.0, 0.6548139945199636) (25.0, 
0.6618568095925998) (26.0, 0.663503977542746) (27.0, 0.66083806677803) (28.0, 0.6629990557996135) (29.0, 0.6683998873866277) (30.0, 0.6648984397080888) (31.0, 0.669578706214518) (32.0, 0.6654962425777016) (33.0, 0.6749729509512238) (34.0, 0.6818469706651584) (35.0, 0.6771993291808562) (36.0, 0.6841698969802391) (37.0, 0.6885233748841056) (38.0, 0.6807809138336006) (39.0, 0.6874126096871727) (40.0, 0.6922103854387927) (41.0, 0.687036770734857) (42.0, 0.6877825195954801) (43.0, 0.6938082106947798) (44.0, 0.6953460530692501) (45.0, 0.6950531287645954) (46.0, 0.6951524141370325) (47.0, 0.6973382994350786) (48.0, 0.6914930995656818) (49.0, 0.6947215532365392) (50.0, 0.6958189424822975) (51.0, 0.6978642511771306) (52.0, 0.696466348371362) (53.0, 0.7006738676191202) (54.0, 0.6982162633926841) (55.0, 0.6989625552814362) (56.0, 0.700445751355786) (57.0, 0.7011865388248453) (58.0, 0.7040575528614621) (59.0, 0.7001947202623636) (60.0, 0.7055893502748461) (61.0, 0.7046465586119333) (62.0, 0.7073084110694532) (63.0, 0.7036066366405402) (64.0, 0.7047139063440171) (65.0, 0.707379155228242) (66.0, 0.7046319994717282) (67.0, 0.7061103813924188) (68.0, 0.707759853977301) (69.0, 0.7001261053125005) (70.0, 0.7045477171074298) (71.0, 0.7042719389110375) (72.0, 0.7024682704396439) (73.0, 0.7051010808773095) (74.0, 0.7065335691532935) (75.0, 0.7052330591751107) (76.0, 0.7012968954171495) (77.0, 0.7066600574025774) (78.0, 0.706633298675242) (79.0, 0.7028829280994057) (80.0, 0.7068536407186623) (81.0, 0.7037813693718707) (82.0, 0.7065505927924821) (83.0, 0.7053901398451693) (84.0, 0.7075539382328201) (85.0, 0.7130248590489746) (86.0, 0.7087316427741849) (87.0, 0.7095614940608663) (88.0, 0.7075232148591263) (89.0, 0.7112732153880439) (90.0, 0.7094338034059238) (91.0, 0.712314285011722) (92.0, 0.712129686523971) (93.0, 0.7106923656740719) (94.0, 0.7113242777657633) (95.0, 0.7128221793130354) (96.0, 0.7099999019651719) (97.0, 0.7114806495877921)
    };
\addplot[color=green, mark=none] coordinates {
(0.0, 0.0781259263571957) (1.0, 0.2051213092742236) (2.0, 0.3023193118792221) (3.0, 0.34003083817855034) (4.0, 0.49784327057420613) (5.0, 0.582979164419174) (6.0, 0.5916705045508864) (7.0, 0.5977061393495671) (8.0, 0.5984389052306339) (9.0, 0.6056397455820322) (10.0, 0.609408590317963) (11.0, 0.6159816756596853) (12.0, 0.6294568411271574) (13.0, 0.6293722717120243) (14.0, 0.6323338026864868) (15.0, 0.6346099855036057) (16.0, 0.638153382915408) (17.0, 0.648418950269642) (18.0, 0.6518943622549265) (19.0, 0.657184877676243) (20.0, 0.659160057945716) (21.0, 0.6626732996258327) (22.0, 0.6589838571601284) (23.0, 0.6662656766415589) (24.0, 0.6689905339348873) (25.0, 0.6718926172628015) (26.0, 0.6731832648676973) (27.0, 0.6754028048701717) (28.0, 0.6725317114216106) (29.0, 0.6795886292410653) (30.0, 0.6782656372604347) (31.0, 0.6781915385958646) (32.0, 0.6759566246375045) (33.0, 0.6823474072726033) (34.0, 0.6877219455298821) (35.0, 0.6846908714447142) (36.0, 0.6908278335233029) (37.0, 0.6883639898336443) (38.0, 0.6813708359805752) (39.0, 0.6942308429376054) (40.0, 0.6941497346245966) (41.0, 0.6905970935469984) (42.0, 0.6913890225170832) (43.0, 0.6963152552162641) (44.0, 0.6966584336293935) (45.0, 0.6968229766648006) (46.0, 0.6976196238486547) (47.0, 0.6970059006904706) (48.0, 0.6994319910401496) (49.0, 0.6937748103608034) (50.0, 0.6975376223363315) (51.0, 0.696870263002034) (52.0, 0.6921874684061902) (53.0, 0.7010163796804485) (54.0, 0.6969228827613603) (55.0, 0.6987341590719472) (56.0, 0.6984234095883188) 
(57.0, 0.6974608013682985) (58.0, 0.6980996024158845) (59.0, 
0.696718344018146) (60.0, 0.6976856770999978) (61.0, 0.697867094499362) (62.0, 0.7015121743891063) (63.0, 0.6981007727775775) (64.0, 0.6979991574811565) (65.0, 0.7028766630929321)
    };
\addplot[color=orange, mark=none] coordinates {
(0.0, 0.07516199929400302) (1.0, 0.20225166100422784) (2.0, 
0.30555528528150033) (3.0, 0.3385569448151113) (4.0, 0.4968880892419172) (5.0, 0.568325435819804) (6.0, 0.5766212761148879) (7.0, 0.5922566966091519) (8.0, 0.5926649292636729) (9.0, 0.5977185638413375) (10.0, 0.6071698844359464) (11.0, 0.6078968103111181) (12.0, 0.6244505287438562) (13.0, 0.6226660953124892) (14.0, 0.6273188841259874) (15.0, 0.6301521066360264) (16.0, 0.6276077737511557) (17.0, 0.6374142439551647) (18.0, 0.6366908585310124) (19.0, 0.6467658979770511) (20.0, 0.6465973209021327) (21.0, 0.6542942686825) (22.0, 0.6497671416440822) (23.0, 0.6558812565423343) (24.0, 0.6575170424975143) (25.0, 0.6618363991054111) (26.0, 0.662332311952019) (27.0, 0.6630716754350804) (28.0, 0.6611583843092995) (29.0, 0.6675727433634546) (30.0, 0.662125740944399) (31.0, 0.6685458920631073) (32.0, 0.6653587444720099) (33.0, 0.6746752935164999) (34.0, 0.6804301659929297) (35.0, 0.6760808163014296) (36.0, 0.6847772377742034) (37.0, 0.6861654313647982) (38.0, 0.6785601896613155) (39.0, 0.6859928333499807) (40.0, 0.6925863772474868) (41.0, 0.6854709328869948) (42.0, 0.6845202322927109) (43.0, 0.6902892739545644) (44.0, 0.6891921525178528) (45.0, 0.6912596173402451) (46.0, 0.6927266047584283) (47.0, 0.6934680295520675) (48.0, 0.6883771834914869) (49.0, 0.6922976178376574) (50.0, 0.6945128334730732) (51.0, 0.6941693098691302) (52.0, 0.6949344558146907) (53.0, 0.6972127034136515) (54.0, 0.692638979214396) (55.0, 0.6946978069610072) (56.0, 0.6943442107423687) (57.0, 0.6980739546846713) (58.0, 0.6985252372260253) (59.0, 0.6945705469307583) (60.0, 0.6992476941102089) (61.0, 0.7015233077357574) (62.0, 0.7049706876892904) (63.0, 0.7016403659071133) (64.0, 0.7015629379813848) (65.0, 0.7061026349045116) 
(66.0, 0.7026649593820272) (67.0, 0.7001405149443626) (68.0, 
0.7054762700130024) (69.0, 0.7003366392303816) (70.0, 0.6987748200357263) (71.0, 0.704967927374922) (72.0, 0.7008151664284668) (73.0, 0.7046265350789824) (74.0, 0.703082515033749) (75.0, 0.7025583121506891) (76.0, 0.6993575432915992) (77.0, 0.7045279567767524) (78.0, 0.703588055996488)
    };
\legend{Bicubic, Nearest-neighbor, Lanczos, Inter-area}
\end{axis}
\end{tikzpicture}
\end{subfigure}
\centering
\begin{subfigure}[b]{\textwidth}
\centering
\begin{tikzpicture}
\begin{axis}[
    xlabel={Epochs},
    ylabel={BCE},
    ytick={0.2, 0.3, 0.4, 0.5, 0.6},
    xmin=0, xmax=100,
    ymin=0.2, ymax=0.6,
    ymajorgrids=true,
    grid style=dashed,
    height=4cm,
    width=9cm
    ]
\addplot[color=blue, mark=none] coordinates {
(0.0, 0.5431148260831833) (1.0, 0.4959254441782832) (2.0, 0.49133785907179117) (3.0, 0.39462773501873016) (4.0, 0.3765806835144758) (5.0, 0.3094360902905464) (6.0, 0.29980515129864216) (7.0, 0.3099737814627588) (8.0, 0.3040202911943197) (9.0, 
0.28797291219234467) (10.0, 0.29765187483280897) (11.0, 0.2949035745114088) (12.0, 0.29095616191625595) (13.0, 0.28456185944378376) (14.0, 0.27933910535648465) (15.0, 0.27690936205908656) (16.0, 0.2737453072331846) (17.0, 0.27816205378621817) (18.0, 0.2637609438970685) (19.0, 0.25611676182597876) (20.0, 
0.26321439584717155) (21.0, 0.2574770199134946) (22.0, 0.24979799566790462) (23.0, 0.26841727970167994) (24.0, 0.26502613676711917) (25.0, 0.24453243473544717) (26.0, 0.25050853146240115) (27.0, 0.2606280446052551) (28.0, 0.2618295238353312) (29.0, 0.25613387720659375) (30.0, 0.2447687853127718) (31.0, 0.25214861892163754) (32.0, 0.23997073993086815) (33.0, 0.23740956094115973) (34.0, 0.23405370861291885) (35.0, 0.25655221845954657) (36.0, 0.24175839452072978) (37.0, 0.24440842028707266) (38.0, 0.2337902751751244) (39.0, 0.24814758822321892) (40.0, 0.23138112016022205) (41.0, 0.23526368895545602) (42.0, 0.23645998444408178) (43.0, 0.2386440229602158) (44.0, 0.2317761075682938) (45.0, 0.23746443074196577) (46.0, 0.24858620576560497) (47.0, 0.26075572380796075) (48.0, 0.24282869789749384) (49.0, 0.24961534002795815) (50.0, 0.22485336381942034) 
(51.0, 0.24816730013117194) (52.0, 0.2263358396012336) (53.0, 0.22952593630179763) (54.0, 0.23529249196872115) (55.0, 0.23316501500084996) (56.0, 0.22333113476634026) (57.0, 0.23848305642604828) (58.0, 0.23764612013474107) (59.0, 0.2399330553598702) (60.0, 0.2340122601017356) (61.0, 0.22868111380375922) 
(62.0, 0.2299410467967391) (63.0, 0.2478275429457426) (64.0, 
0.23376885848119855) (65.0, 0.2408427339978516) (66.0, 0.23154430370777845) (67.0, 0.23636808479204774) (68.0, 0.23331185104325414)
    };
\addplot[color=cyan, mark=none] coordinates {
(0.0, 0.5919985882937908) (1.0, 0.5962236002087593) (2.0, 0.5621740464121103) (3.0, 0.5181903056800365) (4.0, 0.41146075911819935) (5.0, 0.358191542327404) (6.0, 0.3355123568326235) 
(7.0, 0.31720572523772717) (8.0, 0.326081114821136) (9.0, 0.3248825892806053) (10.0, 0.31825452111661434) (11.0, 0.31930592097342014) (12.0, 0.30172027833759785) (13.0, 0.3038948252797127) (14.0, 0.3022042801603675) (15.0, 0.2998025631532073) (16.0, 0.2994705969467759) (17.0, 0.2892692741006613) (18.0, 0.2852422632277012) (19.0, 0.2773164827376604) (20.0, 0.27630473487079144) (21.0, 0.2669798107817769) (22.0, 0.272001382894814) (23.0, 0.2688879622146487) (24.0, 0.2713482240214944) (25.0, 0.2636562008410692) (26.0, 0.260617071762681) (27.0, 0.26832098606973886) (28.0, 0.26393441669642925) (29.0, 0.2565236836671829) (30.0, 0.2645382406190038) (31.0, 0.2561982609331608) (32.0, 0.2632753783836961) (33.0, 0.2524849846959114) (34.0, 0.24814798589795828) (35.0, 0.248075682669878) (36.0, 0.24571033380925655) (37.0, 0.23653512634336948) (38.0, 0.24875320959836245) (39.0, 0.24013720452785492) (40.0, 0.23575259372591972) (41.0, 0.2404219089075923) (42.0, 0.24174141138792038) (43.0, 0.2327156364917755) (44.0, 0.2350675519555807) (45.0, 0.22892808821052313) (46.0, 0.23298692982643843) (47.0, 0.23108868673443794) (48.0, 0.23383859638124704) (49.0, 0.22809523809701204) (50.0, 0.22981545515358448) (51.0, 0.22734692692756653) (52.0, 0.2292936723679304) (53.0, 0.2241621185094118) (54.0, 0.23008527606725693) (55.0, 0.23237009719014168) (56.0, 0.2282525049522519) (57.0, 0.22938275430351496) (58.0, 0.22331217490136623) (59.0, 0.23042484186589718) (60.0, 0.22559050004929304) (61.0, 0.2208197694271803) (62.0, 0.2181796170771122) (63.0, 0.21774187218397856) (64.0, 0.2163865277543664) (65.0, 0.21848888229578733) (66.0, 0.21630681864917278) (67.0, 0.21683748252689838) (68.0, 0.21631054766476154) (69.0, 0.2210534131154418) (70.0, 0.21630607172846794) (71.0, 0.21885761432349682) (72.0, 0.22033047210425138) (73.0, 0.21820516604930162) (74.0, 0.21784122940152884) (75.0, 0.22138409037142992) (76.0, 0.2246285378932953) (77.0, 0.2155496310442686) (78.0, 0.21606117580085993) (79.0, 0.21790700126439333) (80.0, 0.21690980531275272) (81.0, 0.21978462766855955) (82.0, 0.21540951449424028) (83.0, 0.216232149861753) (84.0, 0.21278918720781803) (85.0, 0.2111097676679492) (86.0, 0.21544198412448168) (87.0, 0.21304059214890003) (88.0, 0.2152639701962471) (89.0, 0.21247038803994656) (90.0, 0.21223436761647463) (91.0, 0.21316268481314182) (92.0, 0.21159275993704796) (93.0, 0.21461316477507353) (94.0, 0.2162941535934806) (95.0, 0.21385454200208187) (96.0, 0.21407399605959654) (97.0, 0.2144775101915002)
    };
\addplot[color=green, mark=none] coordinates {
(0.0, 0.5930706933140755) (1.0, 0.5983804613351822) (2.0, 0.5674371682107449) (3.0, 0.5091377310454845) (4.0, 0.4103730656206608) (5.0, 0.34621376916766167) (6.0, 0.3190250461921096) (7.0, 0.3141167052090168) (8.0, 0.3224276388064027) (9.0, 0.31641688849776983) (10.0, 0.31546656135469675) (11.0, 0.3073613056913018) (12.0, 0.2894617738202214) (13.0, 0.29071416705846786) (14.0, 0.28910212498158216) (15.0, 0.2888559540733695) (16.0, 0.28319015447050333) (17.0, 0.27501354459673166) (18.0, 0.27153234742581844) (19.0, 0.2681768052279949) (20.0, 0.26665849052369595) (21.0, 0.26258250419050455) (22.0, 0.26557714957743883) (23.0, 0.2589769996702671) (24.0, 0.25596525706350803) (25.0, 0.2536153383553028) (26.0, 0.2545196609571576) 
(27.0, 0.2520129308104515) (28.0, 0.2541979756206274) (29.0, 
0.24591644946485758) (30.0, 0.24695902038365602) (31.0, 0.2523003416135907) (32.0, 0.25205761939287186) (33.0, 0.2458274168893695) (34.0, 0.2404872141778469) (35.0, 0.2420702464878559) (36.0, 0.2356331516057253) (37.0, 0.23597692977637053) (38.0, 0.25199283845722675) (39.0, 0.23798931390047073) (40.0, 0.23127468023449183) (41.0, 0.2361127771437168) (42.0, 0.23142244573682547) (43.0, 0.2264698576182127) (44.0, 0.22952356468886137) (45.0, 0.22487104777246714) (46.0, 0.22813140414655209) (47.0, 0.22653682343661785) (48.0, 0.2294693822041154) (49.0, 0.2302434490993619) (50.0, 0.22901697643101215) (51.0, 0.23060293402522802) (52.0, 0.23759806901216507) (53.0, 0.2234113859012723) (54.0, 0.23325686901807785) (55.0, 0.2290422674268484) (56.0, 0.2322487961500883) (57.0, 0.22665037587285042) 
(58.0, 0.22963264305144548) (59.0, 0.23167593777179718) (60.0, 0.2331433091312647) (61.0, 0.22780323587357998) (62.0, 0.22575431689620018) (63.0, 0.22602684330195189) (64.0, 0.22688378300517797) (65.0, 0.22624174319207668)
    };
\addplot[color=orange, mark=none] coordinates {
(0.0, 0.5920162796974182) (1.0, 0.596293468028307) (2.0, 0.5624800678342581) (3.0, 0.5175329297780991) (4.0, 0.41139123775064945) (5.0, 0.359032828360796) (6.0, 0.33477003686130047) 
(7.0, 0.317154910415411) (8.0, 0.32607608661055565) (9.0, 0.3252469729632139) (10.0, 0.31863595359027386) (11.0, 0.32100630179047585) (12.0, 0.30195436254143715) (13.0, 0.30419923923909664) (14.0, 0.3011331809684634) (15.0, 0.30025334376841784) (16.0, 0.2999110883101821) (17.0, 0.2911336990073323) (18.0, 0.2876601992174983) (19.0, 0.2778942044824362) (20.0, 0.2761627649888396) (21.0, 0.2671801783144474) (22.0, 0.27078789938241243) (23.0, 0.2696978710591793) (24.0, 0.2682370552793145) (25.0, 0.26420787815004587) (26.0, 0.26316423062235117) (27.0, 0.26523851230740547) (28.0, 0.2656273366883397) (29.0, 0.2575103212147951) (30.0, 0.26739234384149313) (31.0, 0.25685640051960945) (32.0, 0.26258131582289934) (33.0, 0.251796274445951) (34.0, 0.2485528765246272) (35.0, 0.24873329419642687) (36.0, 0.24460276030004025) (37.0, 0.23681770637631416) (38.0, 0.25006781704723835) (39.0, 0.24241926055401564) (40.0, 0.234053835272789) (41.0, 0.23840605188161135) (42.0, 0.24344484321773052) (43.0, 0.23390695173293352) (44.0, 0.23875115904957056) (45.0, 0.23040876630693674) (46.0, 0.2331786686554551) (47.0, 0.22933532763272524) (48.0, 0.23649133555591106) (49.0, 0.23079733829945326) (50.0, 0.23092031944543123) (51.0, 0.2272314950823784) (52.0, 0.2304308507591486) (53.0, 0.2242590319365263) (54.0, 0.23672964982688427) (55.0, 0.23088144324719906) (56.0, 0.2351616658270359) (57.0, 0.22772550769150257) (58.0, 0.227951530367136) (59.0, 0.23218365758657455) (60.0, 0.2287931591272354) (61.0, 0.22297489922493696) (62.0, 0.22235612105578184) (63.0, 0.21862444188445807) (64.0, 0.22032400034368038) (65.0, 0.21676676254719496) (66.0, 0.2167217992246151) (67.0, 0.21948618721216917) (68.0, 0.21700496226549149) (69.0, 0.220673612318933) (70.0, 0.22040299139916897) (71.0, 0.2203191313892603) (72.0, 0.22417333628982306) (73.0, 0.21937969513237476) (74.0, 0.22309754509478807) (75.0, 0.2253790022805333) (76.0, 0.2264071423560381) (77.0, 0.21965876501053572) 
(78.0, 0.22150837257504463)
    };
\end{axis}
\end{tikzpicture}
\end{subfigure}
\caption{The MCC and binary cross-entropy (BCE) scores obtained over the validation set $\ValidationSet$ by the GCNNs trained with patch size equal to 100, and different types of image interpolation. The varying length of the training curves indicates that the early stopping condition has been met.}
\label{fig:val_curves}
\end{figure}

\begin{figure}[ht!]
\scriptsize
\hspace*{-1cm}
\centering
    \hbox{\hspace{0.225cm-2mm}\vspace{-1.52cm}\hspace*{-1cm}
    \begin{subfigure}{0.24\textwidth}
        \includegraphics[width=1\textwidth]{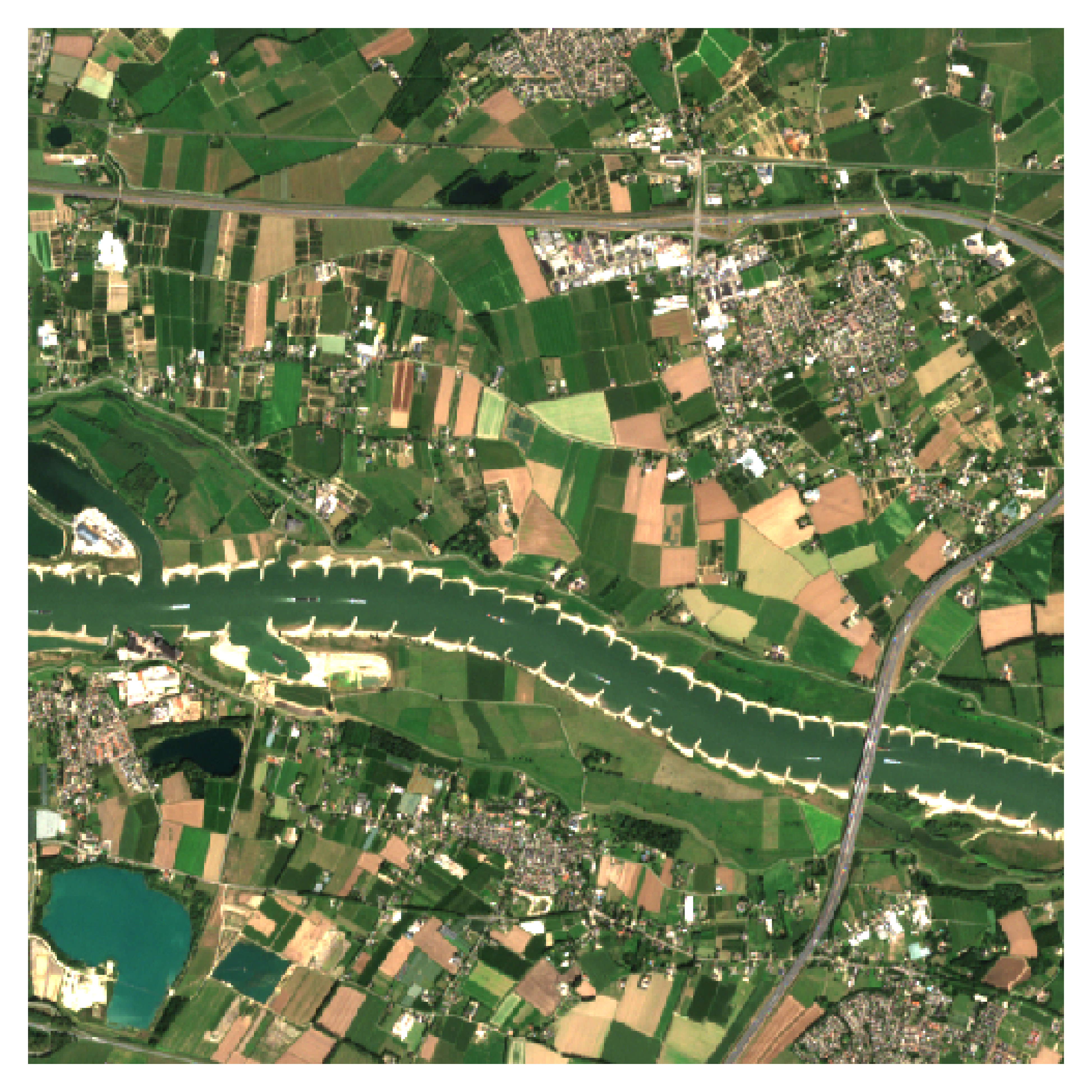}\\[-0.3cm]
        \caption{Example \#1}
    \end{subfigure}
    \begin{subfigure}{0.24\textwidth}
        \begin{tikzpicture}
        [,spy using outlines={circle,cyan,magnification=2,size=2cm, connect spies}]
        \node {\pgfimage[width=1\textwidth]{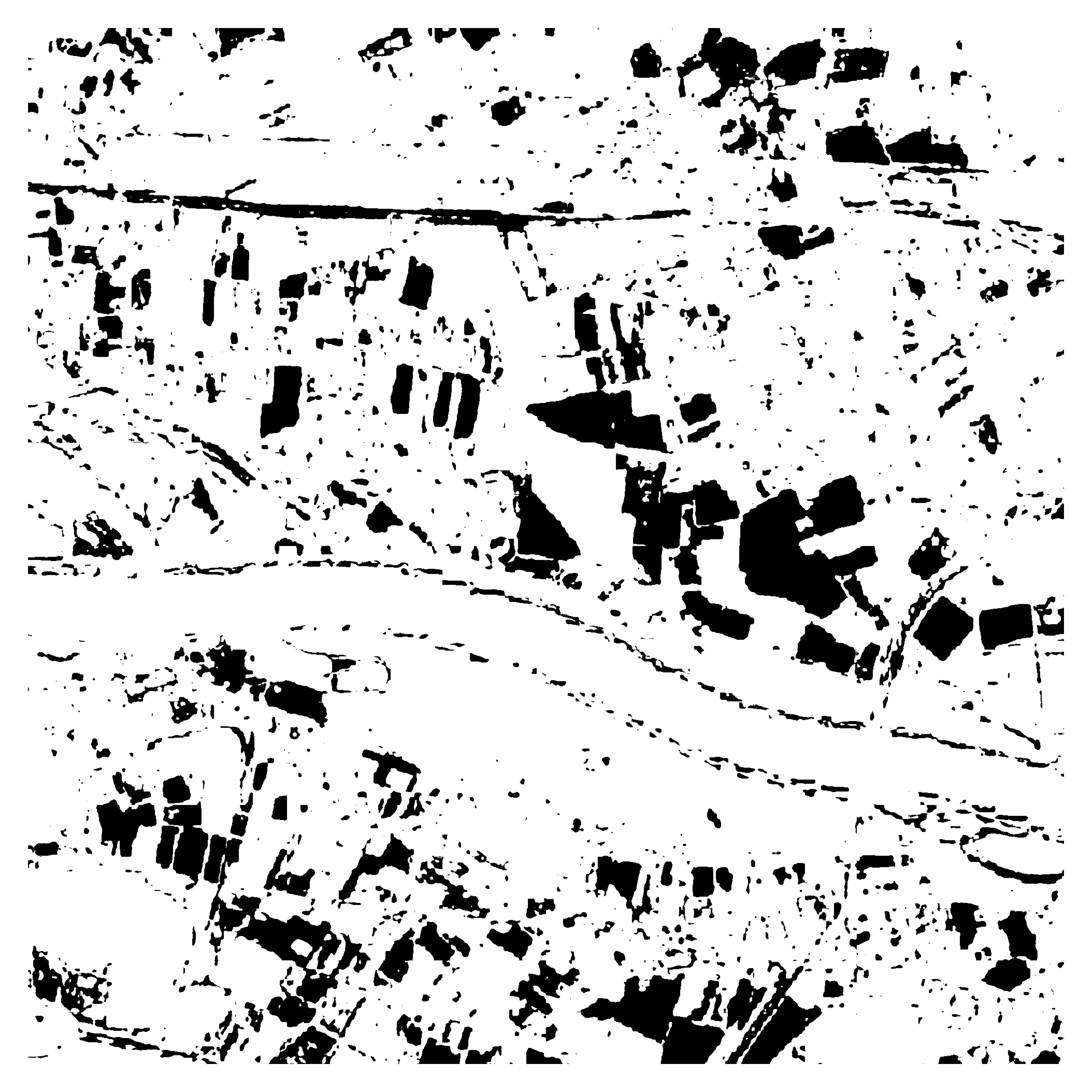}};
        \spy[every spy on node/.append style={thick},every spy in node/.append style={thick}] on (1.0,0.0) in node [left] at (0.1,0.6);
        \end{tikzpicture}
        \caption{Prediction \#1}
    \end{subfigure}}
   
   \hspace*{-2cm}  
     \hbox{\hspace{0.225cm-2mm}\vspace{-0.52cm}
    \begin{subfigure}{0.24\textwidth}
        \includegraphics[width=1\textwidth]{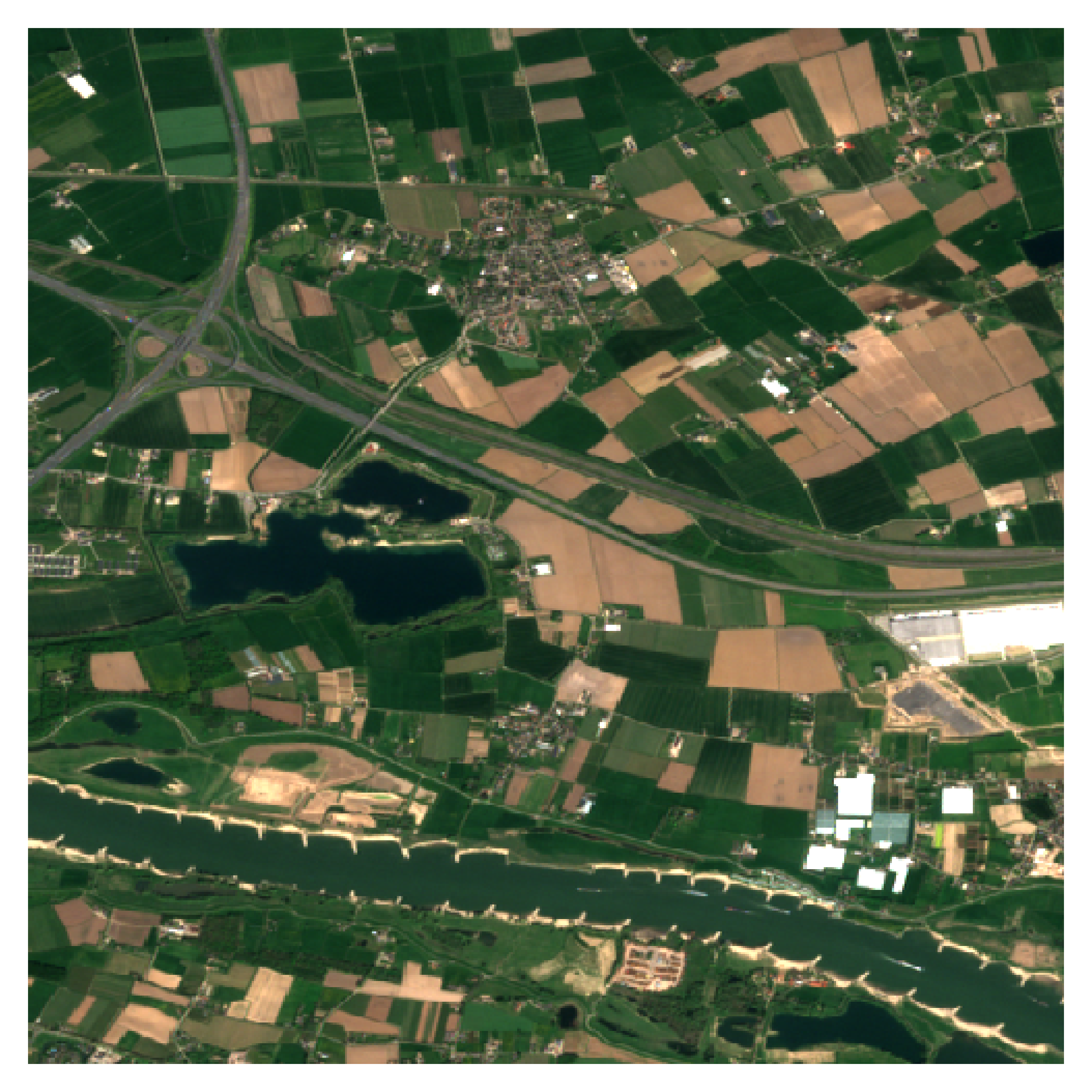}\\[-0.3cm]
        \caption{Example \#2}
    \end{subfigure}
    \begin{subfigure}{0.24\textwidth}
        \begin{tikzpicture}
        [,spy using outlines={circle,cyan,magnification=2,size=2cm, connect spies}]
        \node {\pgfimage[width=1\textwidth]{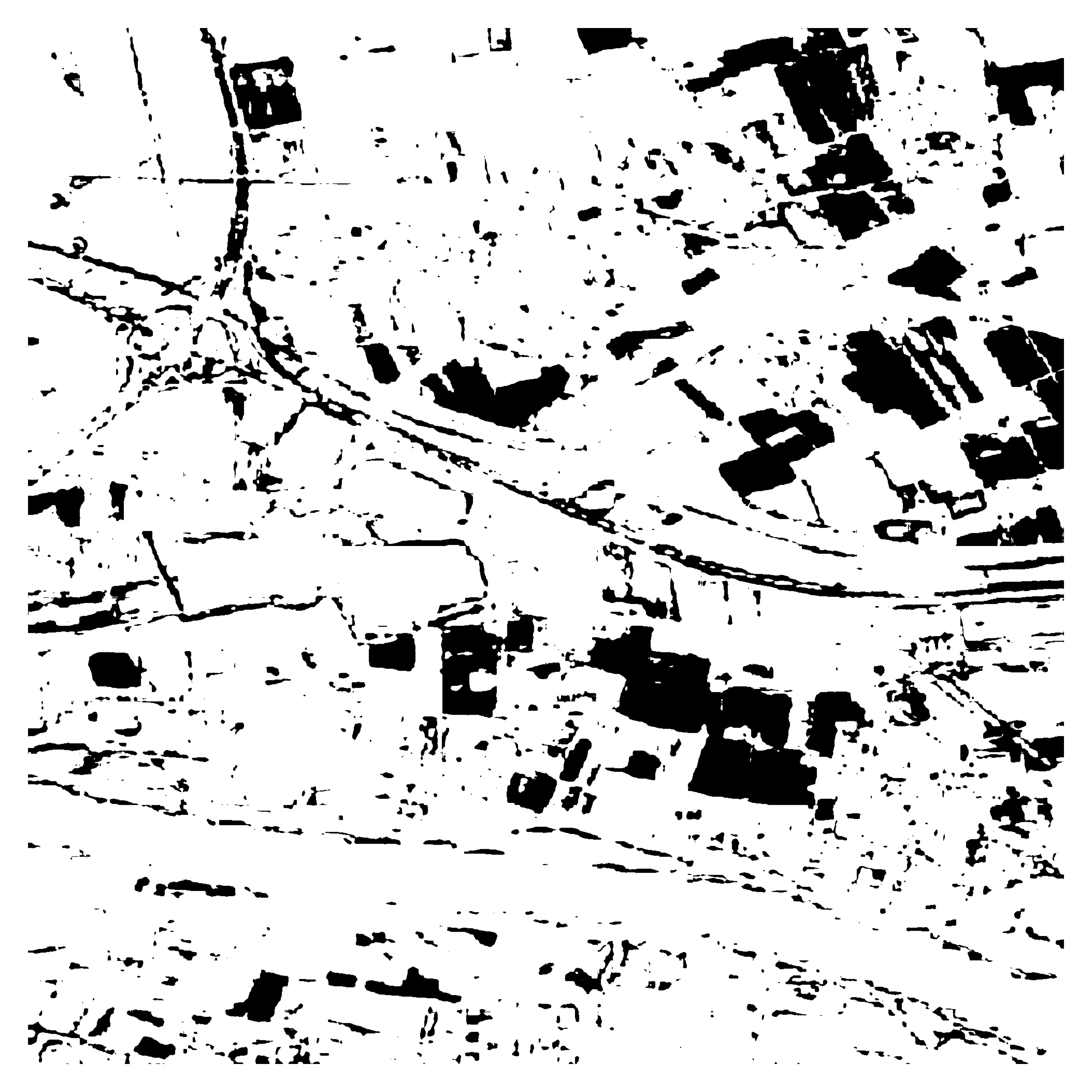}};
        \spy[every spy on node/.append style={thick},every spy in node/.append style={thick}] on (1.4,0.85) in node [left] at (0.1,0.1);
        \end{tikzpicture}
        \caption{Prediction \#2}
    \end{subfigure}}
    
    \hspace*{-2cm} 
    \hbox{\hspace{0.225cm-2mm}\vspace{-0.52cm}
    \begin{subfigure}{0.24\textwidth}
        \includegraphics[width=1\textwidth]{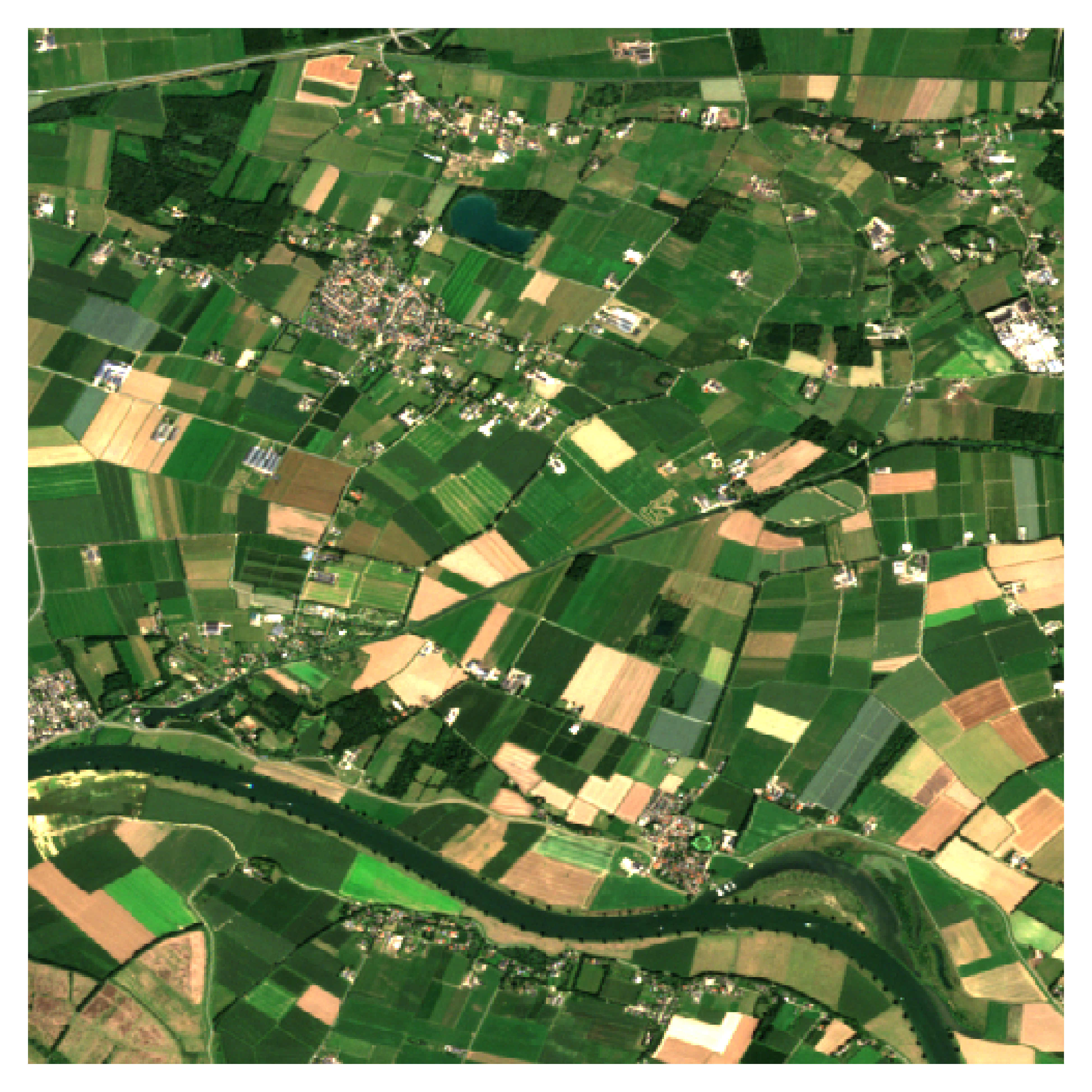}\\[-0.3cm]
        \caption{Example \#3}
    \end{subfigure}
    \begin{subfigure}{0.24\textwidth}
        \begin{tikzpicture}
        [,spy using outlines={circle,cyan,magnification=2,size=2cm, connect spies}]
        \node {\pgfimage[width=1\textwidth]{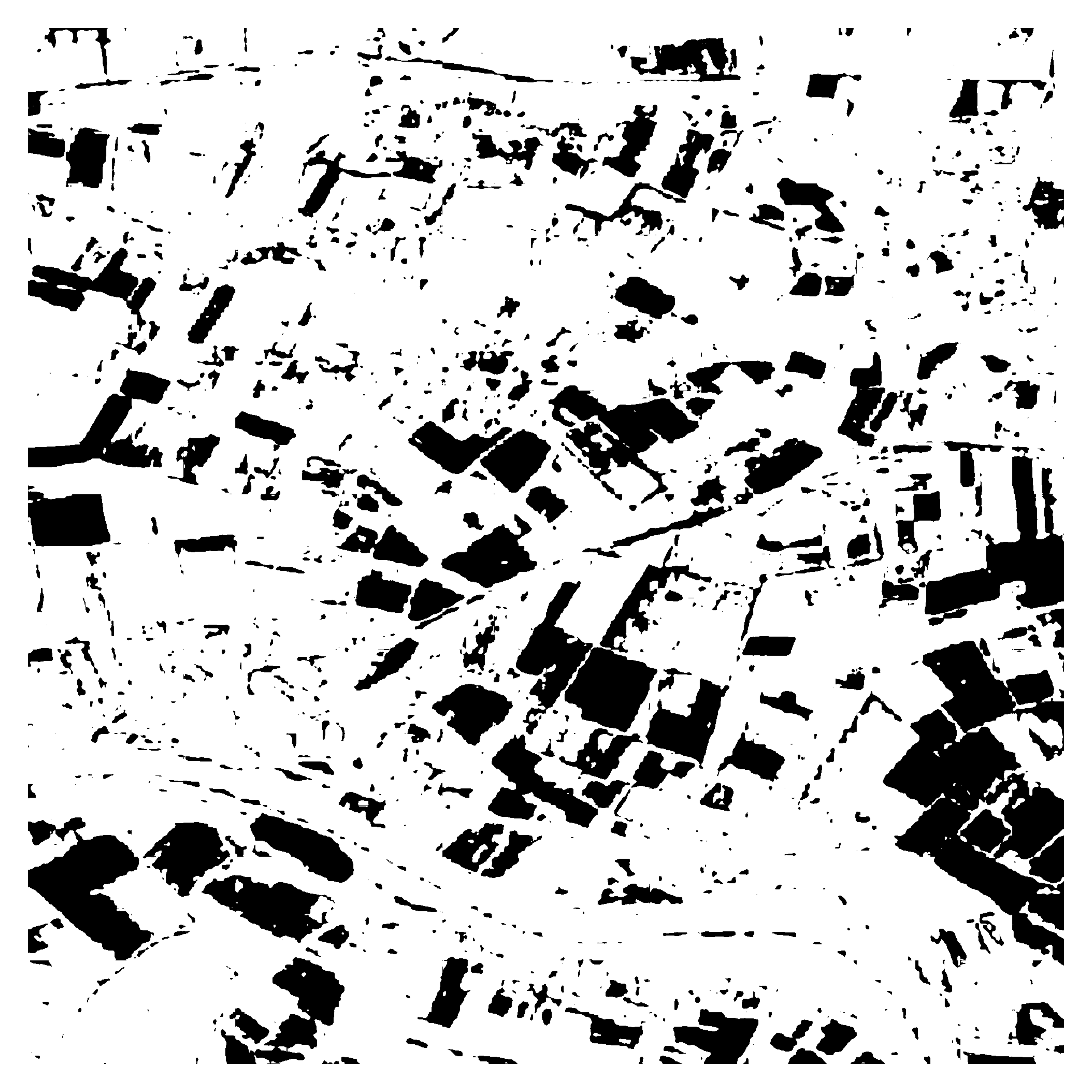}};
        \spy[every spy on node/.append style={thick},every spy in node/.append style={thick}] on (1.4,-0.8) in node [left] at (0.1,0.4);
        \end{tikzpicture}
        \caption{Prediction \#3}
    \end{subfigure}}
    
    \hspace*{-2cm} 
    \hbox{\hspace{0.225cm-2mm}\vspace{-0.52cm}
    \begin{subfigure}{0.24\textwidth}
        \includegraphics[width=1\textwidth]{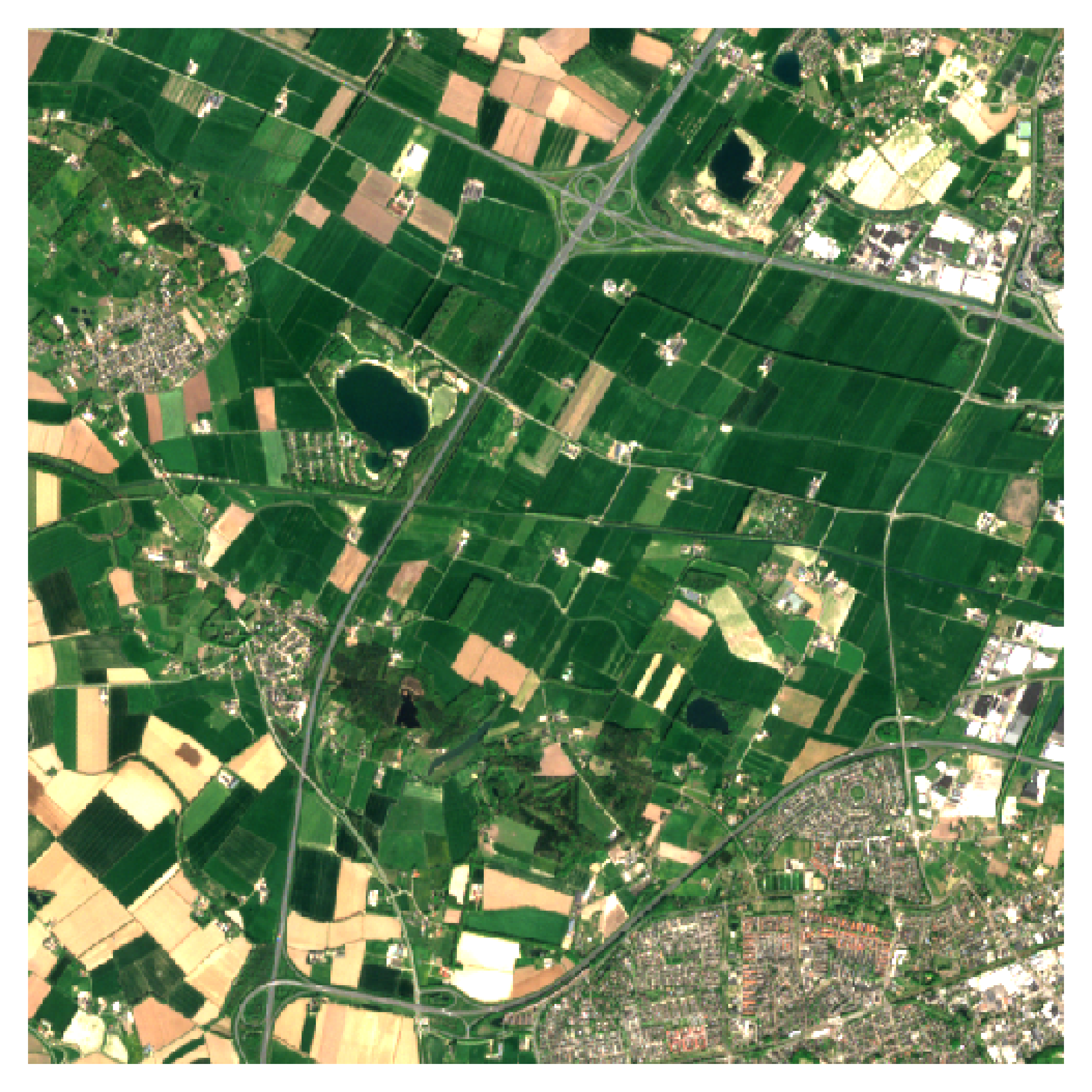}\\[-0.3cm]
        \caption{Example \#4}
    \end{subfigure}
    \begin{subfigure}{0.24\textwidth}
        \begin{tikzpicture}
        [,spy using outlines={circle,cyan,magnification=2,size=2cm, connect spies}]
        \node {\pgfimage[width=1\textwidth]{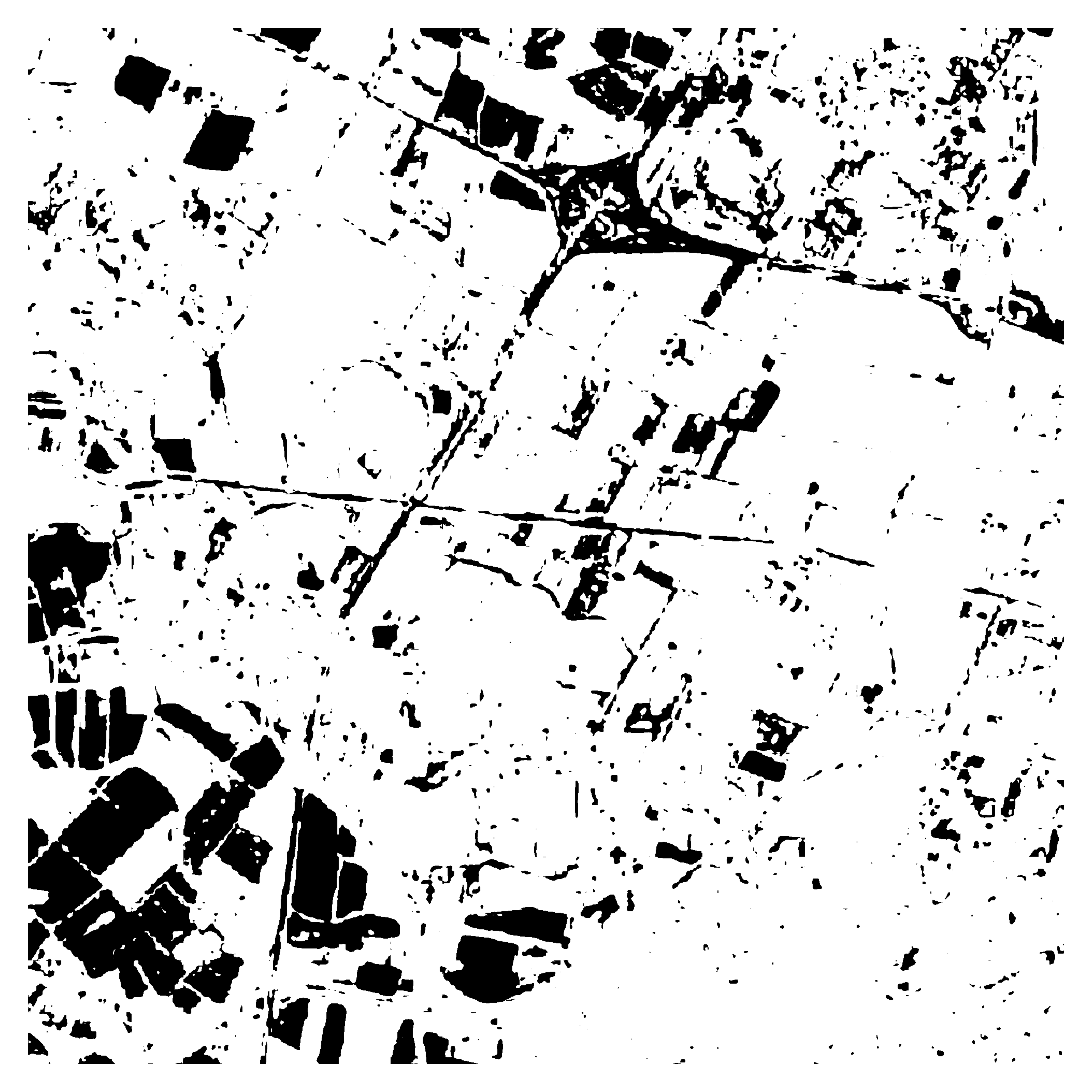}};
        \spy[every spy on node/.append style={thick},every spy in node/.append style={thick}] on (-0.9,-1.25) in node [left] at (1.55,0.4);
        \end{tikzpicture}
        \caption{Prediction \#4}
    \end{subfigure}}
    \caption{Example nine-image Sentinel-2 time-series stacks segmented using the proposed graph neural networks. For each scene, we present an example image from the stack (a, c, e, g), together with the corresponding cultivated land map (some parts of the maps have been zoomed for clarity).}
    \label{fig:examples}
 \end{figure}

We present the training curves elaborated for our graph neural networks coupled with various interpolation techniques used to obtain the 2.5\,m Sentinel-2 images (Figure~\ref{fig:val_curves}), together with the example Sentinel-2 image stacks segmented using our technique (Figure~\ref{fig:examples}).

